\documentclass[10pt,twocolumn,letterpaper]{article}

\usepackage{preprint}

\usepackage{etoolbox}

\usepackage{microtype}
\usepackage{multirow}
\usepackage{amsthm}
\usepackage{bm}
\usepackage{array}
\usepackage{colortbl}%
\usepackage{tabularx}%
\newcolumntype{C}{>{\centering\arraybackslash}X}
\definecolor{pixelred}{HTML}{C9503F}
\definecolor{dctblue}{HTML}{2C6BB3}
\colorlet{pixelband}{pixelred!12}
\colorlet{dctband}{dctblue!12}
\newcolumntype{L}[1]{>{\raggedright\arraybackslash}p{#1}}

\theoremstyle{plain}

\newtheorem{proposition}{Proposition}
\newtheorem{lemma}{Lemma}

\theoremstyle{definition}

\AtEndPreamble{\AtBeginDocument{%
  \crefname{theorem}{Thm.}{Thms.}\Crefname{theorem}{Theorem}{Theorems}%
  \crefname{proposition}{Prop.}{Props.}\Crefname{proposition}{Proposition}{Propositions}%
  \crefname{lemma}{Lem.}{Lems.}\Crefname{lemma}{Lemma}{Lemmas}%
  \crefname{corollary}{Cor.}{Cors.}\Crefname{corollary}{Corollary}{Corollaries}%
  \crefname{definition}{Def.}{Defs.}\Crefname{definition}{Definition}{Definitions}%
  \crefname{equation}{Eq.}{Eqs.}\Crefname{equation}{Equation}{Equations}%
  \crefname{appendix}{App.}{Apps.}\Crefname{appendix}{Appendix}{Appendices}%
  \crefname{subappendix}{App.}{Apps.}\Crefname{subappendix}{Appendix}{Appendices}}}
\AtEndPreamble{\crefname{model}{Model}{Models}\Crefname{model}{Model}{Models}}
\AtEndPreamble{\crefname{assumption}{Assumption}{Assumptions}\Crefname{assumption}{Assumption}{Assumptions}}

\makeatletter
\newcounter{algorithm}
\def\fps@algorithm{tbp}
\def\ftype@algorithm{4}
\def\ext@algorithm{loa}
\def\fnum@algorithm{Algorithm~\thealgorithm}

\makeatother
\AtEndPreamble{\crefname{algorithm}{Alg.}{Algs.}\Crefname{algorithm}{Algorithm}{Algorithms}}
\usepackage{alltt}
\definecolor{codeblue}{rgb}{0.25,0.5,0.5}

\newcommand{\appnumbering}{%
  \setcounter{table}{0}\renewcommand{\thetable}{A\arabic{table}}\def\theHtable{A\arabic{table}}%
  \setcounter{figure}{0}\renewcommand{\thefigure}{A\arabic{figure}}\def\theHfigure{A\arabic{figure}}%
  \setcounter{equation}{0}\renewcommand{\theequation}{A\arabic{equation}}\def\theHequation{A\arabic{equation}}%
  \setcounter{theorem}{0}\renewcommand{\thetheorem}{A\arabic{theorem}}\def\theHtheorem{A\arabic{theorem}}%
  \setcounter{proposition}{0}\renewcommand{\theproposition}{A\arabic{proposition}}%
  \def\theHproposition{A\arabic{proposition}}%
  \setcounter{lemma}{0}\renewcommand{\thelemma}{A\arabic{lemma}}\def\theHlemma{A\arabic{lemma}}%
  \setcounter{corollary}{0}\renewcommand{\thecorollary}{A\arabic{corollary}}\def\theHcorollary{A\arabic{corollary}}%
  \setcounter{definition}{0}\renewcommand{\thedefinition}{A\arabic{definition}}\def\theHdefinition{A\arabic{definition}}%
  \setcounter{model}{0}\renewcommand{\themodel}{A\arabic{model}}\def\theHmodel{A\arabic{model}}%
  \setcounter{assumption}{0}\renewcommand{\theassumption}{A\arabic{assumption}}\def\theHassumption{A\arabic{assumption}}%
  \setcounter{algorithm}{0}\renewcommand{\thealgorithm}{A\arabic{algorithm}}\def\theHalgorithm{A\arabic{algorithm}}}

\newcommand{\ours}{DCT-Rep}

\newcommand{\R}{\mathbb{R}}
\newcommand{\one}{\bm{1}}
\newcommand{\vc}{\bm{c}}\newcommand{\vd}{\bm{d}}\newcommand{\ve}{\bm{e}}\newcommand{\vf}{\bm{f}}\newcommand{\vg}{\bm{g}}\newcommand{\vm}{\bm{m}}\newcommand{\vu}{\bm{u}}\newcommand{\vv}{\bm{v}}\newcommand{\vw}{\bm{w}}\newcommand{\vx}{\bm{x}}\newcommand{\vy}{\bm{y}}\newcommand{\vz}{\bm{z}}
\newcommand{\mC}{\bm{C}}\newcommand{\mD}{\bm{D}}\newcommand{\mE}{\bm{E}}\newcommand{\mF}{\bm{F}}\newcommand{\mH}{\bm{H}}\newcommand{\mI}{\bm{I}}\newcommand{\mL}{\bm{L}}\newcommand{\mQ}{\bm{Q}}\newcommand{\mR}{\bm{R}}\newcommand{\mS}{\bm{S}}\newcommand{\mT}{\bm{T}}\newcommand{\mW}{\bm{W}}\newcommand{\mX}{\bm{X}}\newcommand{\mY}{\bm{Y}}
\newcommand{\vmu}{\bm{\mu}}\newcommand{\vxi}{\bm{\xi}}\newcommand{\vomega}{\bm{\omega}}\newcommand{\vtheta}{\bm{\theta}}\newcommand{\vphi}{\bm{\phi}}
\newcommand{\mSigma}{\bm{\Sigma}}\newcommand{\mTheta}{\bm{\Theta}}
\DeclareMathOperator{\diag}{diag}

\DeclareMathOperator{\sign}{sign}
\newcommand{\E}{\mathbb{E}}

\definecolor{linkblue}{rgb}{0.21,0.49,0.74}
\usepackage[pagebackref,breaklinks,colorlinks,allcolors=linkblue]{hyperref}
\hypersetup{
  pdftitle={Why Adaptive Optimizers Struggle with Large Convolution Kernels},
  pdfauthor={Tianyu Ding, Jiachen Jiang, Jinxin Zhou, Yong Ma, Zhihui Zhu, Luming Liang}
}

\newcommand{\blfootnote}[1]{%
  \begingroup
  \NoHyper
  \renewcommand{\thefootnote}{}%
  \footnotetext{#1}%
  \endNoHyper
  \endgroup
}

\title{Why Adaptive Optimizers Struggle with Large Convolution Kernels}

\author{Tianyu Ding$^{1,*,\dagger}$\quad Jiachen Jiang$^{2,*}$\quad Jinxin Zhou$^{1}$\quad Yong Ma$^{1}$\quad Zhihui Zhu$^{2}$\quad Luming Liang$^{1,\dagger}$\\[4pt]
$^{1}$Microsoft\qquad $^{2}$The Ohio State University}

\begin{document}
\maketitle
\blfootnote{$^*$Equal contribution.\quad $^\dagger$Corresponding authors.}
\begin{abstract}
Large convolution kernels enlarge the receptive fields of ConvNets at a modest computational cost, but they are notoriously difficult to train.
Existing large-kernel networks therefore rely on costly training techniques, such as parallel small-kernel branches with batch normalization.
In this paper, we trace this difficulty to adaptive optimizers such as Adam, which normalize the update of every weight separately.
Since input features rarely average to zero, all weights of a kernel share a common gradient, and these optimizers move the mean of a $k\times k$ kernel $k$ times as far as a single parameter.
We find this \emph{mean amplification} in widely used ConvNets and vision transformers, and show that it can force the maximal learning rate to fall as $1/k^2$.
We further prove that the basis of the discrete cosine transform (DCT) isolates the mean and is the \emph{only} orthonormal basis that diagonalizes every stationary gradient statistic with mirrored boundaries.
We therefore train large kernels through their DCT coefficients and convert them back after training.
The resulting method, \ours{}, requires only a few lines of code and adds no parameters, no inference cost and negligible training overhead.
Extensive experiments on image classification, object detection, semantic segmentation and image restoration show that \ours{} consistently improves large-kernel ConvNets and vision transformers, by $+0.79\%$ top-1 accuracy for a large-kernel ConvNeXt-T and $+0.43\%$ for ViT-S/16 on ImageNet-1K, $+1.4$ box AP on COCO, $+1.2$ mIoU on ADE20K and up to $+0.4$\,dB on image deblurring, where the gain grows with the kernel size.
\end{abstract}

\section{Introduction}
\label{sec:intro}

\begin{figure}[t]
\centering
\includegraphics{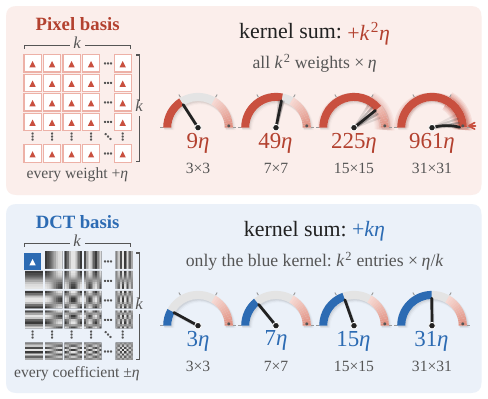}
\vspace{-0.25in}
\caption{\textbf{Mean amplification.} Adaptive optimizers such as Adam move every stored parameter by about the learning rate $\eta$ per step. The logarithmic dials show how far one step moves the sum of a $k\times k$ kernel's weights. Top: in the pixel basis, when input means give all $k^2$ weights one gradient sign, every weight steps by $+\eta$ and the sum moves $k^2\eta$. Bottom: in the DCT basis, only the flat first basis kernel, whose entries are $1/k$, changes the sum; its step of $\eta$ moves it by $k\eta$. The kernel's response to the input mean shifts $k$ times as far in the pixel basis.}
\label{fig:mech}
\vspace{-0.1in}
\end{figure}

Large convolution kernels have regained popularity in modern ConvNets.
A few layers of large kernels capture a wider effective receptive field than a deep stack of small kernels at little extra computation~\cite{luo2016erf,ding2022replknet}, allowing a ConvNet to ``see wide without going deep''~\cite{ding2024unireplknet}.
UniRepLKNet~\cite{ding2024unireplknet} builds on $13\times13$ kernels, and RepLKNet~\cite{ding2022replknet} and SLaK~\cite{liu2023slak} on kernels of $31\times31$ and larger, for image classification, object detection and semantic segmentation.
Image restoration networks such as NAFNet~\cite{chen2022nafnet} can widen their receptive fields in the same way by enlarging their kernels.

Despite these advantages, large kernels are notoriously difficult to train, and naively enlarging them often reduces accuracy~\cite{ding2022replknet,liu2023slak}.
Existing large-kernel networks therefore depend on dedicated training techniques.
RepLKNet, SLaK and UniRepLKNet train small, thin or dilated kernels in parallel branches and place batch normalization (BN)~\cite{ioffe2015bn} after every kernel.
Other works multiply the kernels by spatial masks~\cite{lee2023repux,lee2026rep3d} or initialize them from a trained small-kernel network~\cite{roy2023mednext}.
However, why large kernels need such techniques is not understood, so it is unclear when they are necessary and how to replace them.

\begin{figure}[t]
\centering
\includegraphics{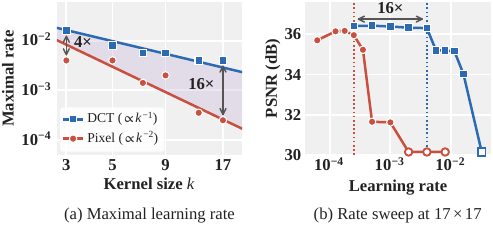}
\vspace{-0.25in}
\caption{\textbf{The maximal learning rate falls as $1/k^2$ in the pixel basis but only as $1/k$ in the DCT basis.} (a)~Largest constant learning rate up to which a super-resolution network trained with Adam ends within 0.5\,dB of its best PSNR, against the kernel size $k$. The lines $c/k^2$ and $c/k$, with $c$ fitted to the measured points, have slopes $-2$ and $-1$ on these log--log axes. The gap between the two bases (shaded) widens with $k$: the DCT basis tolerates a 4 times larger rate than the pixel basis at $3\times3$ and a 16 times larger one at $17\times17$ (arrows). (b)~Validation PSNR at $k=17$ for each constant rate, with the maximal rates from (a) dotted and values below 30\,dB as open markers.}
\label{fig:rate}
\vspace{-0.1in}
\end{figure}

In this paper, we show that the difficulty originates from the optimizer.
Large-kernel networks are usually trained with adaptive optimizers such as Adam~\cite{kingma2015adam}, AdamW~\cite{loshchilov2019adamw}, RMSprop~\cite{tieleman2012rmsprop} and Lion~\cite{chen2023lion}.
These optimizers normalize the update of each parameter, by a running estimate of its gradient magnitude or by its sign, so every parameter moves by about the learning rate $\eta$ per step, however large or small its gradient is.
Since input features are mostly non-negative, as after ReLU-like activations, the gradients of all weights of a kernel share a common component, which often dominates the gradient.
The optimizer then moves all $k^2$ weights of a $k\times k$ kernel by about $\eta$ in the same direction, and the kernel rises as a whole (\cref{fig:mech}, top).
Its mean moves $k$ times as far as it would if it were a single parameter, an effect we call \emph{mean amplification}.
One step thus shifts the kernel's output by $k^2\eta\mu$, where $\mu$ is the input mean.
Because the loss is steep along the kernel's mean, such large shifts destabilize training, and the learning rate must decrease as $1/k^2$ to keep them bounded.
We call the largest learning rate that still trains the \emph{maximal learning rate} (\cref{fig:rate}b).
In a super-resolution network trained with Adam, it falls roughly as $1/k^2$ (\cref{fig:rate}a).

Based on this analysis, we show that giving the kernel's mean a parameter of its own removes mean amplification (\cref{fig:mech}, bottom).
The mean is the simplest of several directions along which a kernel's weight gradients are correlated, such as smooth tilts.
A basis leaves such a direction unamplified only if the direction is a basis kernel.
We prove that the basis of the two-dimensional discrete cosine transform (DCT)~\cite{ahmed1974dct} is the only orthonormal basis that diagonalizes every stationary gradient statistic with mirrored boundaries, so with distinct eigenvalues it contains all of these directions.
On quadratic losses whose Hessian has this form, adaptive optimizers thus train in it as on a perfectly conditioned quadratic.
In this basis, the maximal learning rate falls only roughly as $1/k$ and at $17\times17$ is 16 times that of the pixel basis, against 4 times at $3\times3$ (\cref{fig:rate}a).

We train the large kernels of modern networks in this basis and call this \emph{DCT re-parameterization} (\ours{}).
It takes a few lines of code, and after training the coefficients are converted back into an ordinary kernel, so the deployed network is unchanged.
Earlier works parameterized convolution filters in the same basis, or in the closely related Fourier basis, for compression and faster training~\cite{rippel2015spectral,ulicny2022harmonic,ye2023far}.
Unlike structural re-parameterization~\cite{ding2019acnet,ding2021repvgg,ding2021dbb,zhang2021ecbsr}, \ours{} adds no branches or parameters to training and trains faster, while performing on par with it.

In this paper, we make the following contributions:
\begin{itemize}
\setlength{\itemsep}{1pt}\setlength{\parskip}{0pt}
\item We identify \emph{mean amplification} as {a reason} why large kernels are {hard} to train with adaptive optimizers, and {observe it} in widely used ConvNets and vision transformers.
\item We show that feature means make adaptive optimizers move the mean of a $k\times k$ kernel $k$ times as far as a single parameter and can force the maximal learning rate to fall as $1/k^2$. We prove that the DCT basis isolates the mean and is the only orthonormal basis that diagonalizes every stationary gradient statistic with mirrored boundaries.
\item We train large kernels in the DCT basis as \ours{}, a drop-in re-parameterization without extra parameters or inference cost. Compared with structural re-parameterization, it trains faster {and performs on par}.
\item We conduct extensive experiments on image classification, object detection, semantic segmentation and image restoration with large-kernel ConvNets and vision transformers, and studies across optimizers, normalization layers and kernel sizes. \ours{} consistently improves these networks, with gains that grow with the kernel size.
\end{itemize}

\section{Related work}
\label{sec:related}

Large convolution kernels returned to modern ConvNets with RepLKNet~\cite{ding2022replknet}, which showed that $31\times31$ kernels enlarge the effective receptive field~\cite{luo2016erf} at little extra cost.
RepLKNet trains them with a parallel small kernel that eases optimization and follows every kernel with BN.
Later designs keep these techniques and scale the kernels further.
SLaK~\cite{liu2023slak} exceeds $51\times51$ by decomposing each kernel into two rectangular kernels with BN, UniRepLKNet~\cite{ding2024unireplknet} adds dilated small-kernel branches, and PeLK~\cite{chen2024pelk} shares weights in the periphery of the kernel.
Other works multiply large kernels by fixed~\cite{lee2023repux} or learned~\cite{lee2026rep3d} spatial masks, or build large receptive fields from dilated~\cite{guo2023van} or wavelet-domain~\cite{finder2024wtconv} components.
These works design architectures and training techniques around large kernels.
We instead identify why plain large kernels train poorly under adaptive optimizers, which also explains why BN helps these designs.

Many of these techniques build on structural re-parameterization, which trains extra branches and merges them into a single kernel after training.
ACNet~\cite{ding2019acnet}, RepVGG~\cite{ding2021repvgg}, DBB~\cite{ding2021dbb}, ECB~\cite{zhang2021ecbsr} and ExpandNets~\cite{guo2020expandnets} train additional linear branches or layers in this way, and OREPA~\cite{hu2022orepa} merges them during training to reduce the cost.
Detkov~\etal~\cite{detkov2023spatial} proved that parallel branches act as spatially varying learning rates for optimizers that are linear in the gradient.
RepOptimizer~\cite{ding2023repopt} builds such rates directly into the optimizer, and Rep3D~\cite{lee2026rep3d} derives them under Adam for large three-dimensional kernels.
\ours{} is a re-parameterization in the same sense.
It changes only the basis in which each kernel is stored, so it adds no branches, no parameters and almost no training time.

We find that large kernels need these techniques because of how adaptive optimizers update them.
Our work therefore relates to studies of how parameterization affects optimization.
The parameterization of a network determines the scaling of its optimal learning rate with width~\cite{yang2021mup,everett2024scaling}, and Chen~\etal~\cite{chen2024principled} derive for gradient descent that CNNs with larger kernels should use smaller learning rates.
LeCun~\etal~\cite{lecun1998efficient} advised centering the inputs of a layer, since a non-zero input mean creates a stiff direction.
Adaptive optimizers also depend on the basis in which the parameters are stored.
Adam behaves much like sign descent~\cite{balles2018dissecting,kunstner2023noise}, which is steepest descent in the $\ell_\infty$ geometry of its coordinates~\cite{xie2024implicit}.
Random rotations of the parameters therefore degrade Adam but not SGD~\cite{xie2025adam}.
Some structured rotations help instead, but existing theory does not explain when or why~\cite{zhang2025rotation}.
For convolution kernels, our analysis gives an answer: a good basis amplifies neither the kernel's mean nor any other direction shared by the weight gradients.

Finally, the DCT basis that our analysis leads to is related to earlier frequency-domain parameterizations of filters.
Rippel~\etal~\cite{rippel2015spectral} trained filters through their Fourier coefficients and observed that Adam converged two to five times faster, which they attributed to the sparsity of filter spectra.
Harmonic networks~\cite{ulicny2022harmonic} compose filters from fixed DCT basis functions, and FAR~\cite{ye2023far} trains $3\times3$ kernels as DCT coefficients for overfitted image compression.
GFNet~\cite{rao2021gfnet}, Fourier neural operators~\cite{li2021fno} and NIFF~\cite{grabinski2024nifff} learn large or global filters directly in the frequency domain.
In adaptive filtering, a fixed DCT with per-coefficient power normalization has long sped up LMS on stationary inputs~\cite{narayan1983tdlms,beaufays1995transform}.
\ours{} uses the same DCT parameterization as Harmonic networks and FAR.
Unlike these works, we derive the basis from the gradients of convolution kernels and prove that it is the only orthonormal basis that diagonalizes all of their stationary statistics with mirrored boundaries.
This explains the speed-up that Rippel~\etal found even for $3\times3$ filters, and it points to large kernels as the case where the basis matters most.

\section{Method}
\label{sec:theory}

\subsection{Preliminaries}
\label{sec:setup}

We denote vectors by bold lowercase letters, matrices by bold uppercase letters and scalars by plain letters.
We consider a single $k\times k$ kernel $\vw\in\R^{k^2}$ that connects one input channel to one output channel.
Any such kernel can be written as a linear combination of $k^2$ fixed, orthonormal \emph{basis kernels}.
Its \emph{coefficients} are the parameters that an optimizer stores and updates.
We stack the basis kernels as the rows of an orthogonal $k^2\times k^2$ matrix $\mF$, so that the coefficients are $\vc=\mF\vw$ and the kernel is $\vw=\mF^\top\vc$.
In the standard \emph{pixel basis}, $\mF=\mI$, every basis kernel is a single tap, and the coefficients are the weights themselves.

We split every kernel into its mean and its pattern.
Let the \emph{mean direction} $\ve\in\R^{k^2}$ be the flat kernel whose entries all equal $1/k$, so that $\ve$ has unit norm.
We measure the kernel's mean by its coordinate $\ve^\top\vw$ along this direction and call $\ve^\top\vw$ the \emph{mean coordinate}.
It equals the sum of the weights divided by $k$.
With this scaling, the mean coordinate is a coefficient of every orthonormal basis that contains $\ve$.
We call the rest of $\vw$ the kernel's \emph{pattern}.
It is orthogonal to $\ve$, sums to zero and describes how the weights vary across the window.
In the \emph{DCT basis}, the basis kernels are the products of cosines of the two-dimensional DCT~\cite{ahmed1974dct}.
Its first basis kernel is flat and therefore equal to $\ve$, so its first coefficient is the mean coordinate.
All other basis kernels are zero-sum cosine patterns, so their coefficients change only the pattern.

\emph{Adaptive optimizers} divide the step of each coefficient by a running estimate of its gradient magnitude.
Adam~\cite{kingma2015adam} keeps running averages $\hat m_t$ and $\hat v_t$ of the gradient and its square and updates a coefficient by $-\eta\,\hat m_t/(\sqrt{\hat v_t}+\epsilon)$.
The ratio $\hat m_t/\sqrt{\hat v_t}$ stays near $\pm1$ while the gradient keeps its sign and size, so every coefficient moves by about the learning rate $\eta$ per step, however large its gradient is.
Formally, we call an optimizer adaptive if it is coordinate-wise (each coefficient follows one rule applied to its own gradient history), scale-free (rescaling this history by $a\neq0$ multiplies the steps by $\sign a$) and bounded (no step exceeds a fixed multiple of $\eta$).
Adam, AdamW~\cite{loshchilov2019adamw}, RMSprop~\cite{tieleman2012rmsprop} and AdaGrad~\cite{duchi2011adagrad} are adaptive up to their constant $\epsilon$, and signSGD, Signum~\cite{bernstein2018signsgd} and Lion~\cite{chen2023lion} are exactly adaptive.
SGD is not; its step scales with the gradient.
Note that decoupled weight decay, as in AdamW, shrinks the weights by the same factor in every orthonormal basis, so it neither creates nor removes the mean amplification analyzed next.

\subsection{Mean amplification}
\label{sec:amp}

Mean amplification starts from a gradient that all weights of a kernel share.
Let output channel $o$ of a convolution be $z_o(u)=\sum_{i,p}w_{oi}(p)\,x_i(u+p)$, with input channels $x_i$, positions $u$ and taps $p$, and let $\delta_o=\partial L/\partial z_o$ be the back-propagated gradient.
Splitting each input into its mean $\mu_i$ over positions and the fluctuation $\tilde x_i=x_i-\mu_i$ gives
\begin{equation}
  \frac{\partial L}{\partial w_{oi}(p)}=\underbrace{\mu_i\sum_u\delta_o(u)}_{\substack{\text{common gradient,}\\\text{along }\ve}}+\underbrace{\sum_u\delta_o(u)\,\tilde x_i(u+p)}_{\substack{\text{fluctuation gradient,}\\\text{varies with }p}}.
  \label{eq:split}
\end{equation}
The common gradient is the same at every tap and therefore points along $\ve$, whereas the fluctuation gradient mainly drives the pattern.
Input means are rarely zero, and after ReLU-like activations they are typically positive.
When the common gradient dominates, the gradients of all $k^2$ weights of a kernel have one sign, and Adam moves every weight by about $\eta$ in one direction.
In the pixel basis, the mean coordinate moves by $k^2\eta/k=k\eta$.
In the DCT basis, it is a single coefficient and moves by only $\eta$.

The factor between the two bases is set by how a basis represents $\ve$.
We define the \emph{mean amplification} as the $\ell_1$ norm of the coefficients of $\ve$:
\begin{equation}
  A=\lVert\mF\ve\rVert_1=\textstyle\sum_j\lvert(Fe)_j\rvert.
  \label{def:A}
\end{equation}
It is the key quantity of our analysis, as it bounds how far a step can move the mean coordinate.
Because $\vw=\mF^\top\vc$, the mean coordinate is $\ve^\top\vw=(\mF\ve)^\top\vc=\sum_j(Fe)_jc_j$.
A step that moves every coefficient by at most $\eta$ therefore moves the mean coordinate by at most $\sum_j\lvert(Fe)_j\rvert\eta=A\eta$.
Among all orthonormal bases, the pixel basis has the largest value, $A=\lVert\ve\rVert_1=k$, and only the bases that contain $\ve$, which we call \emph{mean-isolating}, have the smallest, $A=1$.
A random basis spreads $\ve$ over all $k^2$ coefficients and keeps $A$ at about $\sqrt{2/\pi}\,k\approx0.8k$, so a generic change of basis barely helps.
For gradients along $\ve$, we prove that the mean coordinate moves exactly $A$ times as far as a single parameter.\footnote{All proofs and additional results are in the appendix.}

\begin{proposition}[Mean amplification]\label{prop:coherent}
Let every gradient of a kernel up to step $t$ lie along the mean direction, $\vg_s=\gamma_s\ve$, and let an adaptive optimizer with $\epsilon=0$ and no weight decay start from its zero state.
Then in every orthonormal basis, the kernel's mean coordinate moves $A$ times as far as a single parameter $\theta$ that the same optimizer updates with the gradients $\gamma_1,\dots,\gamma_t$, that is, $\ve^\top\Delta\vw_t=A\,\Delta\theta_t$.
\end{proposition}

\begin{table}[t]
\caption{\textbf{Adam amplifies the kernel means of widely used models.} Mean-step ratio: $\mathrm{rms}(\ve^\top\Delta\vw)$ over a layer's kernels for one Adam-like step in the pixel basis, divided by that in the DCT basis, both from the same ImageNet gradients, averaged over layers. It equals $A$ for gradients along $\ve$ and about 1 for gradients with no preferred direction. Norm: normalization after the kernel (LN: layer normalization; --:~none); dw: depthwise.}
\label{tab:probe}
\centering
\footnotesize
\setlength{\tabcolsep}{3pt}
\begin{tabular*}{\linewidth}{@{}l@{\extracolsep{\fill}}l c c c@{}}
\toprule
Model & Kernels & Norm & $A$ & Mean-step ratio \\
\midrule
ConvNeXt-T~\cite{liu2022convnext} & $7\times7$ dw & LN & 7 & 4.3 \\
ConvNeXt V2-T~\cite{woo2023convnextv2} & $7\times7$ dw & LN & 7 & 4.6 \\
ConvFormer-S18~\cite{yu2024metaformer} & $7\times7$ dw & -- & 7 & 4.8 \\
MambaOut-T~\cite{yu2025mambaout} & $7\times7$ dw & -- & 7 & 4.8 \\
RDNet-T~\cite{kim2024rdnet} & $7\times7$ dw & LN & 7 & 4.4 \\
UniRepLKNet-T~\cite{ding2024unireplknet} & $13\times13$ dw & BN & 13 & 4.2 \\
\midrule
DeiT-S~\cite{touvron2021deit} & $16\times16$ patch & LN & 16 & 12.5 \\
CLIP ViT-B/16~\cite{radford2021clip} & $16\times16$ patch & LN & 16 & 13.0 \\
DINOv2 ViT-S/14~\cite{oquab2024dinov2} & $14\times14$ patch & LN & 14 & 11.9 \\
MAE ViT-B/16~\cite{he2022mae} & $16\times16$ patch & LN & 16 & 15.7 \\
\bottomrule
\end{tabular*}

\vspace{-0.1in}
\end{table}

Mean amplification is indeed common in practice, and we observe it in widely used ConvNets and vision transformers.
For the models in \cref{tab:probe}, we compute the gradients of their kernels on ImageNet~\cite{deng2009imagenet} validation images.
\label{sec:probe}From these gradients, we form one Adam step in each basis.
The mean-step ratio measures how many times as far the step of the pixel basis moves the kernel means.
For gradients along $\ve$, \cref{prop:coherent} makes this ratio exactly $k$.
For gradients with no preferred direction, it is about 1.
The closer it is to $k$, the more the gradients concentrate along $\ve$.
In every model, the mean-step ratio lies far above 1, and in the patch embeddings of vision transformers~\cite{dosovitskiy2021vit} it nearly reaches $k$.
Adam thus spends much of every step of these kernels on their means.

\subsection{The price of mean amplification}
\label{sec:rate}\label{sec:means}

Mean amplification has two costs: it lowers the maximal learning rate, and it slows the learning of the patterns.
Through the input means, every step of a kernel's mean shifts the output of the convolution.
The next proposition gives this shift with and without BN after the convolution.

\begin{proposition}[Output shift]\label{prop:centre}
Let every tap see the input means $\vmu=(\mu_i)_i$, as with circular padding.
(i)~In every orthonormal basis, if every kernel that feeds output channel $o$ takes a sign step on its common gradient and $\sum_u\delta_o(u)\neq0$, the spatial average of the channel's output shifts by $\eta kA\lVert\vmu\rVert_1$.
(ii)~With training-mode BN right after the convolution, the common gradient summed over a batch depends on the input means $\vmu^{(b)}$ of its samples $b$ only through their deviations from the batch average $\bar{\vmu}$.
The steps of (i) then shift the spatial average of sample $b$'s centered output by at most $\eta kA\lVert\vmu^{(b)}-\bar{\vmu}\rVert_1$.
\end{proposition}

A negative input mean flips both the common gradient and its effect on the output, so all kernels that feed a channel shift it the same way.
One step therefore shifts the output by $\eta k^2\lVert\vmu\rVert_1$ in the pixel basis ($A=k$) and by $\eta k\lVert\vmu\rVert_1$ in the DCT basis ($A=1$).
For a $31\times31$ kernel, this is 961 times the shift of a $1\times1$ kernel, against 31 times in the DCT basis.
These output shifts limit the learning rate once they exceed what training tolerates in a single step.

We model this limit by a tolerance $T$: training first fails when one step shifts the spatial average of an output channel by more than $T$.
In this model, $T$ depends on the current weights but not on the basis in which they are stored.
\Cref{prop:centre}\,(i) then gives the maximal rate
\begin{equation}
  \tau=\frac{T}{kA\lVert\vmu\rVert_1}.
  \label{eq:taulaw}
\end{equation}
For a $1\times1$ kernel, \cref{eq:taulaw} reduces to $\tau=T/\lVert\vmu\rVert_1$, the classical observation of LeCun~\etal~\cite{lecun1998efficient} that non-zero input means limit the learning rate.
A $k\times k$ kernel tightens this limit by the factor $kA$.
For two runs that pass through the same weights, $T$ and $\vmu$ are common to both.
If they also do not change with $k$, \cref{eq:taulaw} gives
\begin{equation}
  \tau_{\mathrm{pixel}}\propto\frac{1}{k^2},\qquad
  \tau_{\mathrm{DCT}}\propto\frac{1}{k},\qquad
  \frac{\tau_{\mathrm{DCT}}}{\tau_{\mathrm{pixel}}}=k.
  \label{eq:scaling}
\end{equation}
Doubling the kernel width thus divides the maximal rate by four in the pixel basis, but only by two in the DCT basis.
The maximal rates in \cref{fig:rate}a roughly follow these slopes.

Mean amplification also slows the learning of the patterns, and this second cost follows from the first.
To keep training stable, the pixel basis must use a rate about $k$ times smaller.
At this rate, it moves the mean no faster than a mean-isolating basis.
Yet the common gradient still dominates the normalizer of every weight and shrinks the pattern part of each step.
With $r$ the energy of the common gradient relative to the fluctuations, we prove for Gaussian fluctuations and sign or RMS-normalized steps that a mean-isolating basis makes at least $\sqrt{1+r}$ times the expected pattern progress of the slowed pixel basis.
In a super-resolution network, \cref{tab:sr} compares the pixel basis with other bases and with techniques applied to it.
The gain of the DCT basis rises to 0.66\,dB at $k=31$, and slowing the pixel basis by the uniform rate $1/k$ recovers only a third of it.

\begin{table}[t]
\caption{\textbf{No alternative matches the DCT basis at every kernel size.} Urban100 PSNR (dB) of a $\times2$ super-resolution network with four $k\times k$ convolutions. Each entry is the mean of three runs at the best learning rate, and the standard deviation of the DCT basis is at most 0.05\,dB. $A$: mean amplification; bold: column best.}
\label{tab:sr}
\centering
\footnotesize
\setlength{\tabcolsep}{1.6pt}
\begin{tabularx}{\linewidth}{@{}l c CCCCC@{}}
\toprule
& $A$ & $k=7$ & 9 & 13 & 17 & 31 \\
\midrule
\multicolumn{7}{@{}l}{\emph{Bases}} \\
\rowcolor{pixelband}\quad Pixel basis & $k$ & 30.51 & 30.53 & 30.52 & 30.40 & 29.94 \\
\quad Random rotation & ${\approx}0.8k$ & 30.53 & 30.55 & 30.56 & 30.53 & 30.11 \\
\quad Mean-isolating, random & 1 & 30.54 & 30.57 & 30.56 & 30.56 & 30.49 \\
\quad Mean-isolating, pixel-like & 1 & 30.54 & 30.56 & 30.57 & 30.55 & 30.45 \\
\rowcolor{dctband}\quad DCT basis & 1 & \textbf{30.55} & \textbf{30.59} & \textbf{30.60} & \textbf{30.60} & \textbf{30.60} \\
\midrule
\multicolumn{7}{@{}l}{\emph{Pixel basis with}} \\
\quad Gradient centralization & $k$ & 30.31 & 30.05 & 30.45 & 30.34 & 29.40 \\
\quad $3\times3$ branch, no BN & $k{+}9/k$ & 30.54 & 30.56 & 30.51 & 30.45 & 30.26 \\
\quad Fixed distance mask & 2.1--2.9 & 30.52 & 30.56 & 30.47 & {30.55} & 30.57 \\
\quad Per-layer rate $3/k$ & 3 & 30.50 & 30.56 & 30.52 & 30.50 & 30.21 \\
\quad Uniform rate $1/k$ & 1 & 30.37 & 30.52 & 30.29 & 30.50 & 30.16 \\
\midrule
$\Delta$ (DCT $-$ pixel) & & $+$0.04 & $+$0.06 & $+$0.08 & $+$0.20 & $+$0.66 \\
\bottomrule
\end{tabularx}

\label{tab:rivals}
\vspace{-0.1in}
\end{table}

\label{sec:equiv}\label{sec:fields}One could try to avoid these costs in three ways.
First, normalization right after the kernel can remove the output shift.
Training-mode BN subtracts each channel's batch mean, so by \cref{prop:centre}\,(ii) the batch-average input means no longer shift the output.
This explains why large-kernel networks place BN after every kernel and why SLaK found it crucial~\cite{liu2023slak}.
However, the deviations of the input means from their batch average still shift the output.
BN thus reduces what the pixel basis amplifies, but not the amplification itself.
With BN after its kernels, UniRepLKNet~\cite{ding2024unireplknet} in \cref{tab:probe} still shows mean amplification.
Layer normalization over channels (LN)~\cite{ba2016layernorm}, as in ConvNeXt~\cite{liu2022convnext}, removes neither part of the shift, because it is invariant only to shifts shared by all channels at a position.
Second, removing the common gradient before the optimizer falls short.
The kernel means set how each output responds to its input means, so the network must still learn them.
Gradient centralization~\cite{yong2020gc}, which centers each filter's gradient, stays below the DCT basis in \cref{tab:sr} and even below the pixel basis at every kernel size.

Third, one can remove the amplification itself.
The most direct way is an optimizer that ignores the basis.
SGD gives the same deployed kernels in every orthonormal basis, and so do Muon~\cite{jordan2024muon} and Shampoo~\cite{gupta2018shampoo} applied to each layer's kernels as one matrix.
Even adaptive optimizers such as Adam and RMSprop lose the amplification when the norm of a kernel's gradient stays much smaller than their $\epsilon$.
In this {$\epsilon$-dominated regime}, they take SGD-like steps scaled by $\eta/\epsilon$.
Yet SGD trains large kernels worse than every adaptive optimizer (see \cref{fig:pred}b).
Within adaptive optimizers, a \emph{per-tap learning rate}~\cite{detkov2023spatial,lee2026rep3d} rescales the step of each weight by a fixed factor.
Small-kernel branches without BN, the fixed distance mask of RepUX-Net~\cite{lee2023repux} and per-layer learning rates act in this way.
With step factors $\omega_p$ in the pixel basis, a step moves the mean coordinate by at most $\sum_p\omega_p\eta/k$, so $A$ becomes $k$ times their average.\footnote{A $3\times3$ branch receives the same gradients as the nine center weights of the large kernel, and the optimizer gives both the same steps. In the merged kernel, these weights thus have $\omega_p=2$, so $A=(k^2+9)/k=k+9/k$.}
Such factors reduce $A$ only by shrinking the steps, and the patterns still learn slowly.

With an adaptive optimizer, none of these ways removes the amplification without a cost.
They act on the output, the gradient or the steps, not on the basis where the amplification arises.
It is in the basis that we remove it.

\subsection{DCT re-parameterization}
\label{sec:dctbasis}\label{sec:method}

Mean amplification disappears in every mean-isolating basis, in which the mean coordinate is a single coefficient.
The mean is the simplest of the directions along which the gradients of a kernel's weights are correlated.
Since neighboring input features are correlated, the gradients of nearby weights are also correlated along smooth tilts of the kernel.
As with the mean in \cref{def:A}, a step that moves every coefficient by at most $\eta$ moves the kernel along a unit direction $\vy$ by at most $\lVert\mF\vy\rVert_1\eta$, and this factor is 1 only if $\vy$ is a basis kernel.
Bases that isolate the mean but have random or pixel-like pattern directions therefore still amplify the tilts nearly as much as the pixel basis.
In \cref{tab:sr}, they match the DCT basis for small kernels but fall behind it as the kernel grows.
A random rotation does not even isolate the mean and falls behind further.
With distinct eigenvalues, the shared directions are the eigendirections of the gradients' second moment, and a basis amplifies none of them only if it contains them all.

\begin{proposition}[The DCT basis]\label{prop:dctbasis}
Let the second moment of a kernel's gradients be stationary with mirrored boundaries.
The separable DCT basis diagonalizes it, and up to sign flips and reordering it is the \emph{only} orthonormal basis that diagonalizes every second moment of this form.
If the eigenvalues are distinct, it is also the only orthonormal basis that contains all eigendirections, so it amplifies neither the mean nor any other of these directions.
On every positive-definite quadratic loss whose Hessian has this form, adaptive optimizers with $\epsilon=0$ and no weight decay that start from zero state with a sign step, such as Adam, RMSprop, signSGD and Lion, follow in this basis the trajectory of a perfectly conditioned loss.
No other orthonormal basis does so for every such Hessian and initial point.
\end{proposition}

The proposition builds on two known facts: the DCT diagonalizes stationary statistics with mirrored boundaries~\cite{strang1999dct,sanchez1995dct,jain1979sinusoidal}, and a diagonal preconditioner is exact in the eigenbasis of the gradient's second moment~\cite{george2018ekfac}.
Recent optimizers estimate this eigenbasis at extra cost~\cite{vyas2024soap,depavia2025egop} or, for language models, approximate its leading directions by DCT columns~\cite{modoranu2026trion}.
For convolution kernels, \cref{prop:dctbasis} gives this eigenbasis in closed form for statistics mirrored at the window's edges.
On DIV2K patches, the DCT basis keeps the condition number of Adam with frozen second moments within 3--8\% of that of the eigenbasis at $k=3$, 9, 17 and 31.
It thus stays close to the eigenbasis for real image patches.

We train the kernels in the DCT basis and call this re-parameterization \ours{}.
The two-dimensional DCT is separable: with $\mD$ the orthonormal $k\times k$ matrix of the one-dimensional DCT, its $k^2\times k^2$ basis matrix is the Kronecker product $\mF=\mD\otimes\mD$.
\ours{} stores the coefficients $\vc=\mF\vw\in\R^{k^2}$ of each kernel and lets the optimizer update $\vc$ in place of $\vw$.
Because $\mF$ is separable, it never needs to be formed.
In matrix form, the kernel $\mW$ follows from its coefficients $\mC$, and the initial coefficients $\mC_0$ follow from the usual initial kernel $\mW_0$, as
\begin{equation}
  \mW=\mD^\top\mC\mD,\qquad\mC_0=\mD\mW_0\mD^\top.
  \label{eq:dctrep}
\end{equation}
Both bases thus start from the same function; the optimizer, learning rate and weight decay stay unchanged.
After training, we convert back to deploy the original network.

\ours{} adds no parameters, optimizer state or inference cost, and negligible training overhead.
Large kernels gain the most, whereas a $1\times1$ kernel has $A=1$ in every basis.
Since \ours{} changes nothing but the basis, it isolates the effect of the basis and so also serves to validate our analysis in the experiments that follow.

\section{Experiments}
\label{sec:exp}

\begin{figure}[t]
\centering
\includegraphics{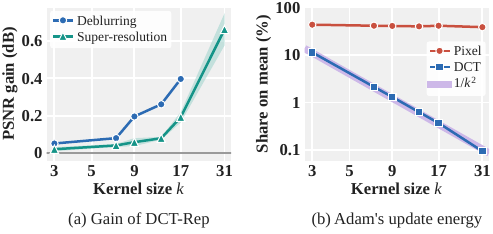}
\vspace{-0.25in}
\caption{\textbf{The gain of \ours{} grows with the kernel size.} (a)~PSNR gain over the pixel basis in the SR network (Urban100, mean of three runs; shaded: $\pm1$ standard error) and in NAFNet deblurring (mean over GoPro and HIDE images). (b)~Share of Adam's update energy on the mean direction in the SR network, against $1/k^2$ for a random direction.}
\label{fig:gain}
\vspace{-0.1in}
\end{figure}

\subsection{Where the gain comes from}
\label{sec:validation}

We first trace the gain of \ours{} to mean amplification in the existing $\times2$ super-resolution (SR) network ECBSR~\cite{zhang2021ecbsr}, whose $3\times3$ body convolutions we enlarge to $k\times k$.
We train each configuration with Adam on DIV2K~\cite{agustsson2017div2k} at its best learning rate on validation images and report its PSNR on the Urban100 test set~\cite{huang2015urban100}.

We measure the gain $\Delta$ of \ours{} as its result minus that of the pixel basis for the same network and kernel size.
As \cref{fig:gain}a shows, the gain grows with the kernel size in the SR network and reaches $+0.66$\,dB at $31\times31$, where both bases train on larger patches.
NAFNet deblurring (\cref{sec:published}) shows the same growth up to $17\times17$.
In the SR network, mean amplification explains this growth, as \cref{fig:gain}b shows.
In the pixel basis, Adam spends 39--43\% of its update energy, the squared length of its step, on the mean direction at every $k$ from 3 to 31.
In the DCT basis, the mean is one of $k^2$ coefficients that each move by about $\eta$, so it receives only about $1/k^2$ of this energy, as a random direction would.
The pixel basis thus gives the mean about $4\times$ the energy that the DCT basis gives it at $3\times3$ and about $400\times$ at $31\times31$.

In \cref{tab:sr}, isolating the mean accounts for most of the gain {at large kernels}.
At $k=31$, a mean-isolating basis with random pattern directions reaches 0.55\,dB of the 0.66\,dB that the DCT basis gains, whereas a random rotation, which still amplifies the mean ($A\approx0.8k$), reaches only 0.17\,dB.
The smooth pattern directions of the DCT basis add the rest, at least 0.11\,dB over both other mean-isolating bases.

\begin{figure}[t]
\centering
\includegraphics{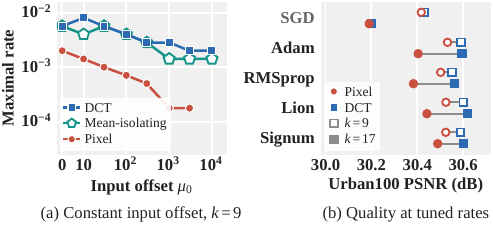}
\vspace{-0.25in}
\caption{\textbf{The input mean and the optimizer control the gap.} (a)~Maximal rate against a constant input offset $\mu_0$ at $k=9$. (b)~Urban100 PSNR of each optimizer at its tuned rate, in the pixel basis (circles) and the DCT basis (squares).}
\label{fig:pred}
\vspace{-0.1in}
\end{figure}

If mean amplification causes the gap between the bases, the gap should depend on the input mean, on the optimizer and on BN after the kernel.
All three predictions hold.
First, a larger input mean widens the gap in maximal rate, as \cref{fig:pred}a shows.
We add a constant $\mu_0$ to the input of every body convolution, padding included, and subtract $\mu_0$ times the initial kernel sum from its bias.
This leaves the initial network unchanged and adds a pure common gradient to every kernel, which enters every weight's normalizer in the pixel basis but only the mean coefficient's in the DCT basis.
An offset of $\mu_0=1000$ lowers the maximal rate of the pixel basis by $11\times$ and its best PSNR by 1.5\,dB, and by 2.8\,dB at $\mu_0=3000$.
The DCT basis still trains at $\mu_0=10^4$, at a third of its maximal rate and with a 0.6\,dB lower PSNR.
A mean-isolating basis with pixel-like pattern directions loses about as little, so isolating the mean suffices for this robustness.

Second, the gain appears under all four adaptive optimizers and vanishes under SGD, as \cref{fig:pred}b shows.
We tune each optimizer's learning rate separately in each basis and report the mean of three runs.
At $k=9$, the DCT basis gains 0.06, 0.05, 0.08 and 0.06\,dB on Urban100 under Adam, RMSprop~\cite{tieleman2012rmsprop}, Lion~\cite{chen2023lion} and Signum~\cite{bernstein2018signsgd}, and at $k=17$ it gains 0.20, 0.18, 0.18 and 0.11\,dB.
Under SGD, the two bases differ by only 0.01\,dB at both kernel sizes, as its basis invariance predicts.
Yet SGD is no remedy: it avoids mean amplification but ends 0.1 and 0.2\,dB below even Adam in the pixel basis at $k=9$ and 17, so giving up adaptive step sizes costs more than the amplification.

\begin{table}[t]
\centering
\caption{\textbf{BN lifts the maximal rate of the pixel basis far above practical rates; LN does not.} Maximal rate $\tau$ and best DIV2K validation PSNR of the SR network, each setting after every body convolution; with BN, the $3\times3$ branch has BN on both paths.}
\label{tab:norm}
\scriptsize
\setlength{\tabcolsep}{1.5pt}
\begin{tabular*}{\linewidth}{@{}l@{\extracolsep{\fill}}cccc cccc@{}}
\toprule
& \multicolumn{4}{c}{$k=9$} & \multicolumn{4}{c}{$k=17$} \\
\cmidrule(lr){2-5}\cmidrule(l){6-9}
& \multicolumn{2}{c}{$\tau$ ($10^{-3}$)} & \multicolumn{2}{c}{PSNR (dB)} & \multicolumn{2}{c}{$\tau$ ($10^{-3}$)} & \multicolumn{2}{c}{PSNR (dB)} \\
\cmidrule(lr){2-3}\cmidrule(lr){4-5}\cmidrule(lr){6-7}\cmidrule(l){8-9}
Setting & Pixel & DCT & Pixel & DCT & Pixel & DCT & Pixel & DCT \\
\midrule
No normalization & 2 & 5.7 & 36.38 & 36.46 & 0.25 & 4 & 36.14 & 36.42 \\
Centered inputs & 1 & 5.7 & 36.41 & 36.45 & 0.35 & 5.7 & 35.99 & 36.44 \\
\midrule
BN & 128 & 256 & 36.41 & 36.46 & 11.3 & 362 & 36.35 & 36.43 \\
LN & 2 & 1024 & 36.33 & 36.40 & 4 & 512 & 36.24 & 36.37 \\
\midrule
$3\times3$ branch, no BN & 2 & 4 & 36.41 & 36.49 & 0.25 & 2.8 & 36.26 & 36.43 \\
$3\times3$ branch with BN & 91 & 128 & 36.39 & 36.48 & 256 & 362 & 36.48 & 36.51 \\
\bottomrule
\end{tabular*}

\vspace{-0.1in}
\end{table}

\begin{table}[t]
\centering
\caption{\textbf{\ours{} improves large-kernel ConvNets and vision transformers across tasks.} ConvNeXt-T with 31--29--27--13 kernels; detection and segmentation fine-tune its $4\times10^{-3}$ backbones. ViT-S/16 trains only its patch embedding as DCT coefficients. Tuned rate: best on held-out training images. Each entry is the mean of three runs{, with a standard deviation of at most 0.2 points}. Bold: better of the two.}
\label{tab:transfer}
\footnotesize
\setlength{\tabcolsep}{2pt}
\begin{tabular*}{\linewidth}{@{}l@{\extracolsep{\fill}}ccc@{}}
\toprule
Task, network and metric & Baseline & \ours{} & $\Delta$ \\
\midrule
\multicolumn{4}{@{}l}{\emph{Image classification, ImageNet-1K top-1 (\%)}} \\
\quad ConvNeXt-T, each at its tuned rate & 80.43 & \textbf{81.22} & $+0.79$ \\
\quad ConvNeXt-T, both at $4\times10^{-3}$ & 80.43 & \textbf{80.96} & $+0.53$ \\
\quad ViT-S/16, each at its tuned rate & 76.71 & \textbf{77.14} & $+0.43$ \\
\quad ViT-S/16, both at $10^{-3}$ & 76.71 & \textbf{77.09} & $+0.38$ \\
\midrule
\multicolumn{4}{@{}l}{\emph{Object detection, ConvNeXt-T backbone}} \\
\quad COCO box AP, Cascade Mask R-CNN & 46.2 & \textbf{47.6} & $+1.4$ \\
\quad COCO mask AP, Cascade Mask R-CNN & 40.3 & \textbf{41.5} & $+1.2$ \\
\quad VOC mAP, Faster R-CNN & 81.5 & \textbf{82.4} & $+0.9$ \\
\midrule
\multicolumn{4}{@{}l}{\emph{Semantic segmentation, ConvNeXt-T backbone}} \\
\quad ADE20K mIoU, UperNet & 43.54 & \textbf{44.75} & $+1.21$ \\
\bottomrule
\end{tabular*}

\vspace{-0.1in}
\end{table}

Third, BN after each kernel lifts the maximal rate of the pixel basis far above the rates used in training, whereas LN does not, as \cref{tab:norm} shows.
Training-mode BN~\cite{ioffe2015bn} removes the batch-shared part of the common gradient (\cref{prop:centre}), and its normalization of scale lets both bases train at much larger rates.
Under BN, the maximal rate of the pixel basis rises by $64\times$ at $k=9$ and by $45\times$ at $k=17$, whereas under LN it stays unchanged at $k=9$ and rises only by $16\times$ at $k=17$.
The amplification itself remains: under BN, the DCT basis still trains at $2\times$ and $32\times$ the maximal rate of the pixel basis and gains 0.05 and 0.08\,dB in PSNR at $k=9$ and 17.
LN normalizes the scale as well and also lifts the DCT basis above $500\times10^{-3}$, but it leaves the common gradient intact, so the pixel basis gains little.
Under LN, the DCT basis thus trains at $512\times$ and $128\times$ the maximal rate of the pixel basis and leads by 0.07 and 0.13\,dB.
Centering the input of every body convolution by its running mean, as LeCun~\etal~\cite{lecun1998efficient} advise, still leaves the DCT basis $6\times$ and $16\times$ the maximal rate of the pixel basis.
Centering by a running mean removes only the average input mean: each image's own deviation from it and the zero-padded borders still feed a common gradient.
A $3\times3$ branch raises the maximal rate only through its BN.
At $k=17$, the branch without BN leaves the maximal rate of the pixel basis unchanged, whereas with BN it raises this rate by about $1000\times$ and reaches a PSNR above the unnormalized DCT basis.

These experiments confirm the mechanism of mean amplification: input means lead to a common gradient, adaptive optimizers amplify it along the kernel's mean, and the resulting mean step limits the maximal rate.
The gain of \ours{} thus comes from its basis, which isolates the kernel's mean and nearly decorrelates the gradients of its other coefficients.

\subsection{Impact across vision tasks}
\label{sec:published}

The gain carries over to widely used networks across image classification, object detection, semantic segmentation and image deblurring.
We train only the large kernels as DCT coefficients and use the same recipe in both bases.

For image classification, we train ConvNeXt-T~\cite{liu2022convnext} on ImageNet-1K~\cite{deng2009imagenet} with 31--29--27--13 depthwise kernels, under the 120-epoch recipe of SLaK~\cite{liu2023slak} with AdamW and without BN after the kernels, as in ConvNeXt.
With the learning rate tuned for each basis on held-out training images, \ours{} raises its top-1 accuracy from 80.43\% to 81.22\%, and at the baseline's rate it still gains 0.53 points (\cref{tab:transfer}).
A vision transformer gains as well, since it starts with a large-kernel patch embedding (\cref{tab:probe}), where Chen~\etal~\cite{chen2021mocov3} located the instability of ViT training.
We train only the patch embedding of ViT-S/16~\cite{dosovitskiy2021vit} as DCT coefficients, one $16\times16$ DCT per color channel, and use the 90-epoch recipe of Beyer~\etal~\cite{beyer2022better}.
Its top-1 accuracy rises from 76.71\% to 77.14\%, and by 0.38 points at the baseline's rate.

For object detection and semantic segmentation, we fine-tune the baseline and \ours{} backbones trained at $4\times10^{-3}$ under SLaK's protocols and keep the \ours{} backbone in the DCT basis.
Both tasks benefit, as \cref{tab:transfer} shows.
On COCO~\cite{lin2014coco}, Cascade Mask R-CNN~\cite{cai2018cascade,he2017maskrcnn} with the 1$\times$ schedule gains 1.4 box AP and 1.2 mask AP.
On VOC~\cite{everingham2010pascal}, Faster R-CNN~\cite{ren2015faster} gains 0.9 mAP.
On ADE20K~\cite{zhou2017ade20k}, UperNet~\cite{xiao2018upernet} gains 1.21 mIoU.

\begin{table}[t]
\centering
\caption{\textbf{The deblurring gain grows with the kernel size.} Gain $\Delta$ (dB) of \ours{} over the pixel basis for NAFNet with $k\times k$ kernels, trained on GoPro. Images better: share of all 3,136 test images. Each entry is the mean of three runs{, with a standard deviation of at most 0.03\,dB}.}
\label{tab:restore}
\scriptsize
\setlength{\tabcolsep}{1.5pt}
\begin{tabular*}{\linewidth}{@{}l@{\extracolsep{\fill}}ccccc@{}}
\toprule
Test set & $k=3$ & 7 & 9 & 13 & 17 \\
\midrule
GoPro & $+0.05$ & $+0.03$ & $+0.16$ & $+0.29$ & $+0.41$ \\
HIDE & $+0.05$ & $+0.11$ & $+0.21$ & $+0.25$ & $+0.39$ \\
\midrule
All 3,136 images & $+0.05$ & $+0.08$ & $+0.20$ & $+0.26$ & $+0.40$ \\
Images better & 72\% & 69\% & 90\% & 93\% & 96\% \\
\bottomrule
\end{tabular*}

\end{table}

For image deblurring, we enlarge the $3\times3$ depthwise kernels of NAFNet~\cite{chen2022nafnet} of width 32 to $k\times k$, with $k=3$, 7, 9, 13 and 17, and name each network by its kernel size, as in NAFNet-k13.
We train on GoPro~\cite{nah2017gopro} with its official recipe and test on GoPro and HIDE~\cite{shen2019hide} with the test-time local converter (TLC)~\cite{chu2022tlc}.
As in the SR network, the gain grows with the kernel size (\cref{tab:restore}).
Averaged over all 3,136 test images, it rises from $+0.05$\,dB at $3\times3$ to $+0.40$\,dB at $17\times17$, where \ours{} is better on 96\% of the images.
The gain comes from optimization rather than regularization, since NAFNet-k13 with \ours{} also fits its GoPro training images better than the pixel basis at every evaluation.
We stop at $17\times17$, about half the width of the $32\times32$ training feature maps at 1/8 resolution{, where 28 of NAFNet's 36 blocks operate}. Across these four tasks, DCT-Rep changes only how the large kernels are trained, so each network keeps its architecture, recipe and inference cost.

\begin{table}[t]
\centering
\caption{\textbf{\ours{} matches structural re-parameterizations at almost no extra training cost.} Urban100 PSNR of the SR network, each method at its best rate on 100 DIV2K validation images, mean of three runs with a standard deviation of at most 0.02\,dB for \ours{}; training cost relative to the baseline at $k=9$ and 17 (time per update, trained parameters, memory). Bold: column best. --:~the method needs a kernel larger than $3\times3$.}
\label{tab:reparam}
\scriptsize
\setlength{\tabcolsep}{1.5pt}
\begin{tabularx}{\linewidth}{@{}l CCCC ccc@{}}
\toprule
& \multicolumn{4}{c}{PSNR (dB) at kernel size $k$} & \multicolumn{3}{c}{Training cost ($\times$)} \\
\cmidrule(lr){2-5}\cmidrule(l){6-8}
Method & 3 & 7 & 13 & 17 & Time & Param. & Memory \\
\midrule
\rowcolor{pixelband}Baseline (pixel basis) & 30.23 & {30.51} & {30.52} & 30.40 & 1.0 & 1.00 & 1.0 \\
\midrule
\multicolumn{8}{@{}l}{\emph{Branches with BN}} \\
\quad ACNet~\cite{ding2019acnet} & 30.17 & 30.50 & 30.56 & 30.58 & 1.7 & 1.15 & 1.9 \\
\quad RepVGG-style~\cite{ding2021repvgg} & 30.15 & 30.47 & 30.55 & 30.50 & 1.4 & 1.01 & 1.6 \\
\quad DBB~\cite{ding2021dbb} & 30.16 & 30.52 & 30.55 & 30.54 & 4.0 & 2.02 & 3.8 \\
\quad Small kernel~\cite{ding2022replknet,liu2023slak} & -- & 30.50 & 30.56 & 30.58 & 1.5 & 1.05 & 1.6 \\
\quad Dilated~\cite{ding2024unireplknet} & -- & 30.50 & 30.58 & 30.59 & 3.5 & 1.60 & 2.9 \\
\midrule
\multicolumn{8}{@{}l}{\emph{Branches without BN}} \\
\quad ECB~\cite{zhang2021ecbsr} & 30.22 & 30.48 & 30.45 & 30.35 & 1.4--1.5 & 3.0 & 1.1 \\
\quad Larger expansion & 30.22 & 30.51 & {30.04} & 30.53 & 1.5--1.7 & 50--55 & 1.5--2.7 \\
\midrule
\multicolumn{8}{@{}l}{\emph{Spatial mask}} \\
\quad Learned mask~\cite{lee2026rep3d} & -- & 30.49 & 30.58 & 30.59 & 1.2--1.4 & 1.00 & 1.0 \\
\midrule
\rowcolor{dctband}\ours{} (DCT basis) & \textbf{30.25} & \textbf{30.55} & \textbf{30.60} & \textbf{30.60} & 1.0--1.1 & 1.00 & 1.0 \\
\bottomrule
\end{tabularx}

\end{table}

\subsection{Relation to structural re-parameterization}
\label{sec:reparam}

\ours{} is a re-parameterization method, so we compare it with structural re-parameterizations in the SR network at $k=3$, 7, 13 and 17 (\cref{tab:reparam}).
We also build a larger linear expansion that widens each convolution into parallel branches, as in ECB~\cite{zhang2021ecbsr}, and deepens each branch into $1\times1$--$k\times k$--$1\times1$ chains, as in ExpandNets~\cite{guo2020expandnets}.
This expansion trains about 50 times as many parameters, yet it stays below \ours{} at every kernel size.

\ours{} is on par with the best of them at every kernel size, and its gain over the baseline grows from 0.02\,dB at $3\times3$ to 0.20\,dB at $17\times17$.
From $7\times7$ on, the branch methods with BN come close, since the BN on the large kernel's own path removes its batch-shared common gradient.
At $17\times17$, ACNet, the small-kernel and the dilated branches come within 0.02\,dB of \ours{}, and so does the learned mask, whose rates adapt during training.
ECB, whose branches carry no BN, falls further behind \ours{} as the kernel grows and {stays below the baseline}.
These methods deploy the same network, but \ours{} adds no parameters, negligible memory and at most 6\% to the step time, against 1.4--4.0 times for branches.

\section{Conclusion}
\label{sec:conclusion}

We identified mean amplification as {a reason} why large convolution kernels are hard to train with adaptive optimizers: these optimizers move the mean of a $k\times k$ kernel $k$ times as far as a single parameter, and feature means make the kernel gradients share this direction.
{This also explains why existing large-kernel networks place BN after every kernel.}
We proved that the DCT basis isolates the mean and is the only orthonormal basis that diagonalizes every stationary gradient statistic with mirrored boundaries.
Training large kernels in this basis improves ConvNeXt, NAFNet and ViT across classification, detection, segmentation and deblurring at no inference cost{, and it performs on par with structural re-parameterization while training faster and adding no parameters}.
Beyond convolution kernels, we expect the same principle to help adaptive optimizers wherever the gradients of many weights share known directions.

{
    \small
    \bibliographystyle{ieeenat_fullname}
    \bibliography{refs}

\begin{thebibliography}{81}
\providecommand{\natexlab}[1]{#1}
\providecommand{\url}[1]{\texttt{#1}}
\expandafter\ifx\csname urlstyle\endcsname\relax
  \providecommand{\doi}[1]{doi: #1}\else
  \providecommand{\doi}{doi: \begingroup \urlstyle{rm}\Url}\fi

\bibitem[Agustsson and Timofte(2017)]{agustsson2017div2k}
Eirikur Agustsson and Radu Timofte.
\newblock {NTIRE} 2017 challenge on single image super-resolution: Dataset and
  study.
\newblock In \emph{IEEE Conference on Computer Vision and Pattern Recognition
  Workshops (CVPRW)}, pages 126--135, 2017.

\bibitem[Ahmed et~al.(1974)Ahmed, Natarajan, and Rao]{ahmed1974dct}
N. Ahmed, T. Natarajan, and K.~R. Rao.
\newblock Discrete cosine transform.
\newblock \emph{IEEE Transactions on Computers}, C-23\penalty0 (1):\penalty0
  90--93, 1974.

\bibitem[Ba et~al.(2016)Ba, Kiros, and Hinton]{ba2016layernorm}
Jimmy~Lei Ba, Jamie~Ryan Kiros, and Geoffrey~E. Hinton.
\newblock Layer normalization.
\newblock \emph{arXiv preprint arXiv:1607.06450}, 2016.

\bibitem[Balles and Hennig(2018)]{balles2018dissecting}
Lukas Balles and Philipp Hennig.
\newblock Dissecting {Adam}: The sign, magnitude and variance of stochastic
  gradients.
\newblock In \emph{International Conference on Machine Learning (ICML)}, pages
  404--413, 2018.

\bibitem[Beaufays(1995)]{beaufays1995transform}
Fran{\c{c}}oise Beaufays.
\newblock Transform-domain adaptive filters: An analytical approach.
\newblock \emph{IEEE Transactions on Signal Processing}, 43\penalty0
  (2):\penalty0 422--431, 1995.

\bibitem[Bernstein et~al.(2018)Bernstein, Wang, Azizzadenesheli, and
  Anandkumar]{bernstein2018signsgd}
Jeremy Bernstein, Yu-Xiang Wang, Kamyar Azizzadenesheli, and Animashree
  Anandkumar.
\newblock {signSGD}: Compressed optimisation for non-convex problems.
\newblock In \emph{International Conference on Machine Learning (ICML)}, pages
  560--569, 2018.

\bibitem[Beyer et~al.(2022)Beyer, Zhai, and Kolesnikov]{beyer2022better}
Lucas Beyer, Xiaohua Zhai, and Alexander Kolesnikov.
\newblock Better plain {ViT} baselines for {ImageNet-1k}.
\newblock \emph{arXiv preprint arXiv:2205.01580}, 2022.

\bibitem[Cai and Vasconcelos(2018)]{cai2018cascade}
Zhaowei Cai and Nuno Vasconcelos.
\newblock Cascade {R-CNN}: Delving into high quality object detection.
\newblock In \emph{IEEE/CVF Conference on Computer Vision and Pattern
  Recognition (CVPR)}, pages 6154--6162, 2018.

\bibitem[Chen et~al.(2024{\natexlab{a}})Chen, Chu, Ren, Zhao, and
  Huang]{chen2024pelk}
Honghao Chen, Xiangxiang Chu, Yongjian Ren, Xin Zhao, and Kaiqi Huang.
\newblock {PeLK}: Parameter-efficient large kernel {ConvNets} with peripheral
  convolution.
\newblock In \emph{IEEE/CVF Conference on Computer Vision and Pattern
  Recognition (CVPR)}, pages 5557--5567, 2024{\natexlab{a}}.

\bibitem[Chen et~al.(2022)Chen, Chu, Zhang, and Sun]{chen2022nafnet}
Liangyu Chen, Xiaojie Chu, Xiangyu Zhang, and Jian Sun.
\newblock Simple baselines for image restoration.
\newblock In \emph{European Conference on Computer Vision (ECCV)}, pages
  17--33, 2022.

\bibitem[Chen et~al.(2024{\natexlab{b}})Chen, Wu, Wang, and
  Hanin]{chen2024principled}
Wuyang Chen, Junru Wu, Zhangyang Wang, and Boris Hanin.
\newblock Principled architecture-aware scaling of hyperparameters.
\newblock In \emph{International Conference on Learning Representations
  (ICLR)}, 2024{\natexlab{b}}.

\bibitem[Chen et~al.(2021)Chen, Xie, and He]{chen2021mocov3}
Xinlei Chen, Saining Xie, and Kaiming He.
\newblock An empirical study of training self-supervised vision transformers.
\newblock In \emph{IEEE/CVF International Conference on Computer Vision
  (ICCV)}, pages 9640--9649, 2021.

\bibitem[Chen et~al.(2023)Chen, Liang, Huang, Real, Wang, Pham,
  et~al.]{chen2023lion}
Xiangning Chen, Chen Liang, Da Huang, Esteban Real, Kaiyuan Wang, Hieu Pham,
  et~al.
\newblock Symbolic discovery of optimization algorithms.
\newblock In \emph{Advances in Neural Information Processing Systems
  (NeurIPS)}, pages 49205--49233, 2023.

\bibitem[Chu et~al.(2022)Chu, Chen, Chen, and Lu]{chu2022tlc}
Xiaojie Chu, Liangyu Chen, Chengpeng Chen, and Xin Lu.
\newblock Improving image restoration by revisiting global information
  aggregation.
\newblock In \emph{European Conference on Computer Vision (ECCV)}, pages
  53--71, 2022.

\bibitem[Deng et~al.(2009)Deng, Dong, Socher, Li, Li, and
  Fei-Fei]{deng2009imagenet}
Jia Deng, Wei Dong, Richard Socher, Li-Jia Li, Kai Li, and Li Fei-Fei.
\newblock {ImageNet}: A large-scale hierarchical image database.
\newblock In \emph{IEEE Conference on Computer Vision and Pattern Recognition
  (CVPR)}, pages 248--255, 2009.

\bibitem[DePavia et~al.(2025)DePavia, Cruzado, Liang, Charisopoulos, and
  Willett]{depavia2025egop}
Adela DePavia, Jose Cruzado, Jiayou Liang, Vasileios Charisopoulos, and Rebecca
  Willett.
\newblock Faster adaptive optimization via expected gradient outer product
  reparameterization.
\newblock \emph{arXiv preprint arXiv:2502.01594}, 2025.

\bibitem[Detkov et~al.(2023)Detkov, Salameh, Qharabagh, Zhang, Lu, Jui, and
  Niu]{detkov2023spatial}
Alexander Detkov, Mohammad Salameh, Muhammad~Fetrat Qharabagh, Jialin Zhang,
  Wei Lu, Shangling Jui, and Di Niu.
\newblock Reparameterization through spatial gradient scaling.
\newblock In \emph{International Conference on Learning Representations
  (ICLR)}, 2023.

\bibitem[Ding et~al.(2019)Ding, Guo, Ding, and Han]{ding2019acnet}
Xiaohan Ding, Yuchen Guo, Guiguang Ding, and Jungong Han.
\newblock {ACNet}: Strengthening the kernel skeletons for powerful {CNN} via
  asymmetric convolution blocks.
\newblock In \emph{IEEE/CVF International Conference on Computer Vision
  (ICCV)}, pages 1911--1920, 2019.

\bibitem[Ding et~al.(2021{\natexlab{a}})Ding, Zhang, Han, and
  Ding]{ding2021dbb}
Xiaohan Ding, Xiangyu Zhang, Jungong Han, and Guiguang Ding.
\newblock Diverse branch block: Building a convolution as an inception-like
  unit.
\newblock In \emph{IEEE/CVF Conference on Computer Vision and Pattern
  Recognition (CVPR)}, pages 10886--10895, 2021{\natexlab{a}}.

\bibitem[Ding et~al.(2021{\natexlab{b}})Ding, Zhang, Ma, Han, Ding, and
  Sun]{ding2021repvgg}
Xiaohan Ding, Xiangyu Zhang, Ningning Ma, Jungong Han, Guiguang Ding, and Jian
  Sun.
\newblock {RepVGG}: Making {VGG}-style {ConvNets} great again.
\newblock In \emph{IEEE/CVF Conference on Computer Vision and Pattern
  Recognition (CVPR)}, pages 13733--13742, 2021{\natexlab{b}}.

\bibitem[Ding et~al.(2022)Ding, Zhang, Han, and Ding]{ding2022replknet}
Xiaohan Ding, Xiangyu Zhang, Jungong Han, and Guiguang Ding.
\newblock Scaling up your kernels to 31x31: Revisiting large kernel design in
  {CNNs}.
\newblock In \emph{IEEE/CVF Conference on Computer Vision and Pattern
  Recognition (CVPR)}, pages 11963--11975, 2022.

\bibitem[Ding et~al.(2023)Ding, Chen, Zhang, Huang, Han, and
  Ding]{ding2023repopt}
Xiaohan Ding, Honghao Chen, Xiangyu Zhang, Kaiqi Huang, Jungong Han, and
  Guiguang Ding.
\newblock Re-parameterizing your optimizers rather than architectures.
\newblock In \emph{International Conference on Learning Representations
  (ICLR)}, 2023.

\bibitem[Ding et~al.(2024)Ding, Zhang, Ge, Zhao, Song, Yue, and
  Shan]{ding2024unireplknet}
Xiaohan Ding, Yiyuan Zhang, Yixiao Ge, Sijie Zhao, Lin Song, Xiangyu Yue, and
  Ying Shan.
\newblock {UniRepLKNet}: A universal perception large-kernel {ConvNet} for
  audio, video, point cloud, time-series and image recognition.
\newblock In \emph{IEEE/CVF Conference on Computer Vision and Pattern
  Recognition (CVPR)}, pages 5513--5524, 2024.

\bibitem[Dosovitskiy et~al.(2021)Dosovitskiy, Beyer, Kolesnikov, Weissenborn,
  Zhai, Unterthiner, Dehghani, Minderer, Heigold, Gelly, Uszkoreit, and
  Houlsby]{dosovitskiy2021vit}
Alexey Dosovitskiy, Lucas Beyer, Alexander Kolesnikov, Dirk Weissenborn,
  Xiaohua Zhai, Thomas Unterthiner, Mostafa Dehghani, Matthias Minderer, Georg
  Heigold, Sylvain Gelly, Jakob Uszkoreit, and Neil Houlsby.
\newblock An image is worth 16x16 words: Transformers for image recognition at
  scale.
\newblock In \emph{International Conference on Learning Representations
  (ICLR)}, 2021.

\bibitem[Duchi et~al.(2011)Duchi, Hazan, and Singer]{duchi2011adagrad}
John Duchi, Elad Hazan, and Yoram Singer.
\newblock Adaptive subgradient methods for online learning and stochastic
  optimization.
\newblock \emph{Journal of Machine Learning Research}, 12\penalty0
  (61):\penalty0 2121--2159, 2011.

\bibitem[Everett et~al.(2024)Everett, Xiao, Wortsman, Alemi, Novak, Liu,
  et~al.]{everett2024scaling}
Katie~E. Everett, Lechao Xiao, Mitchell Wortsman, Alexander~A. Alemi, Roman
  Novak, Peter~J. Liu, et~al.
\newblock Scaling exponents across parameterizations and optimizers.
\newblock In \emph{International Conference on Machine Learning (ICML)}, pages
  12666--12700, 2024.

\bibitem[Everingham et~al.(2010)Everingham, {Van Gool}, Williams, Winn, and
  Zisserman]{everingham2010pascal}
Mark Everingham, Luc {Van Gool}, Christopher K.~I. Williams, John Winn, and
  Andrew Zisserman.
\newblock The {Pascal} visual object classes ({VOC}) challenge.
\newblock \emph{International Journal of Computer Vision}, 88\penalty0
  (2):\penalty0 303--338, 2010.

\bibitem[Finder et~al.(2024)Finder, Amoyal, Treister, and
  Freifeld]{finder2024wtconv}
Shahaf~E. Finder, Roy Amoyal, Eran Treister, and Oren Freifeld.
\newblock Wavelet convolutions for large receptive fields.
\newblock In \emph{European Conference on Computer Vision (ECCV)}, pages
  363--380, 2024.

\bibitem[George et~al.(2018)George, Laurent, Bouthillier, Ballas, and
  Vincent]{george2018ekfac}
Thomas George, C{\'e}sar Laurent, Xavier Bouthillier, Nicolas Ballas, and
  Pascal Vincent.
\newblock Fast approximate natural gradient descent in a {Kronecker} factored
  eigenbasis.
\newblock In \emph{Advances in Neural Information Processing Systems
  (NeurIPS)}, 2018.

\bibitem[Grabinski et~al.(2024)Grabinski, Keuper, and
  Keuper]{grabinski2024nifff}
Julia Grabinski, Janis Keuper, and Margret Keuper.
\newblock As large as it gets -- studying infinitely large convolutions via
  neural implicit frequency filters.
\newblock \emph{Transactions on Machine Learning Research}, 2024.

\bibitem[Guo et~al.(2023)Guo, Lu, Liu, Cheng, and Hu]{guo2023van}
Meng-Hao Guo, Cheng-Ze Lu, Zheng-Ning Liu, Ming-Ming Cheng, and Shi-Min Hu.
\newblock Visual attention network.
\newblock \emph{Computational Visual Media}, 9\penalty0 (4):\penalty0 733--752,
  2023.

\bibitem[Guo et~al.(2020)Guo, Alvarez, and Salzmann]{guo2020expandnets}
Shuxuan Guo, Jose~M. Alvarez, and Mathieu Salzmann.
\newblock {ExpandNets}: Linear over-parameterization to train compact
  convolutional networks.
\newblock In \emph{Advances in Neural Information Processing Systems
  (NeurIPS)}, pages 1298--1310, 2020.

\bibitem[Gupta et~al.(2018)Gupta, Koren, and Singer]{gupta2018shampoo}
Vineet Gupta, Tomer Koren, and Yoram Singer.
\newblock Shampoo: Preconditioned stochastic tensor optimization.
\newblock In \emph{International Conference on Machine Learning (ICML)}, pages
  1842--1850, 2018.

\bibitem[He et~al.(2017)He, Gkioxari, Doll{\'a}r, and Girshick]{he2017maskrcnn}
Kaiming He, Georgia Gkioxari, Piotr Doll{\'a}r, and Ross Girshick.
\newblock Mask {R-CNN}.
\newblock In \emph{IEEE International Conference on Computer Vision (ICCV)},
  pages 2961--2969, 2017.

\bibitem[He et~al.(2022)He, Chen, Xie, Li, Doll{\'a}r, and Girshick]{he2022mae}
Kaiming He, Xinlei Chen, Saining Xie, Yanghao Li, Piotr Doll{\'a}r, and Ross
  Girshick.
\newblock Masked autoencoders are scalable vision learners.
\newblock In \emph{IEEE/CVF Conference on Computer Vision and Pattern
  Recognition (CVPR)}, pages 16000--16009, 2022.

\bibitem[Hu et~al.(2022)Hu, Feng, Hua, Lai, Huang, Gong, and Hua]{hu2022orepa}
Mu Hu, Junyi Feng, Jiashen Hua, Baisheng Lai, Jianqiang Huang, Xiaojin Gong,
  and Xian-Sheng Hua.
\newblock Online convolutional re-parameterization.
\newblock In \emph{IEEE/CVF Conference on Computer Vision and Pattern
  Recognition (CVPR)}, pages 568--577, 2022.

\bibitem[Huang et~al.(2015)Huang, Singh, and Ahuja]{huang2015urban100}
Jia-Bin Huang, Abhishek Singh, and Narendra Ahuja.
\newblock Single image super-resolution from transformed self-exemplars.
\newblock In \emph{IEEE Conference on Computer Vision and Pattern Recognition
  (CVPR)}, pages 5197--5206, 2015.

\bibitem[Ioffe and Szegedy(2015)]{ioffe2015bn}
Sergey Ioffe and Christian Szegedy.
\newblock Batch normalization: Accelerating deep network training by reducing
  internal covariate shift.
\newblock In \emph{International Conference on Machine Learning (ICML)}, pages
  448--456, 2015.

\bibitem[Jain(1979)]{jain1979sinusoidal}
Anil~K. Jain.
\newblock A sinusoidal family of unitary transforms.
\newblock \emph{IEEE Transactions on Pattern Analysis and Machine
  Intelligence}, PAMI-1\penalty0 (4):\penalty0 356--365, 1979.

\bibitem[Jordan et~al.(2024)Jordan, Jin, Boza, You, Cesista, Newhouse, and
  Bernstein]{jordan2024muon}
Keller Jordan, Yuchen Jin, Vlado Boza, Jiacheng You, Franz Cesista, Laker
  Newhouse, and Jeremy Bernstein.
\newblock Muon: An optimizer for hidden layers in neural networks.
\newblock Blog post, \url{https://kellerjordan.github.io/posts/muon/}, 2024.

\bibitem[Kim et~al.(2024)Kim, Heo, and Han]{kim2024rdnet}
Donghyun Kim, Byeongho Heo, and Dongyoon Han.
\newblock {DenseNets} reloaded: Paradigm shift beyond {ResNets} and {ViTs}.
\newblock In \emph{European Conference on Computer Vision (ECCV)}, pages
  395--415, 2024.

\bibitem[Kingma and Ba(2015)]{kingma2015adam}
Diederik~P. Kingma and Jimmy Ba.
\newblock Adam: A method for stochastic optimization.
\newblock In \emph{International Conference on Learning Representations
  (ICLR)}, 2015.

\bibitem[Kunstner et~al.(2023)Kunstner, Chen, Lavington, and
  Schmidt]{kunstner2023noise}
Frederik Kunstner, Jacques Chen, Jonathan~Wilder Lavington, and Mark Schmidt.
\newblock Noise is not the main factor behind the gap between {SGD} and {Adam}
  on transformers, but sign descent might be.
\newblock In \emph{International Conference on Learning Representations
  (ICLR)}, 2023.

\bibitem[LeCun et~al.(1998)LeCun, Bottou, Orr, and
  M{\"u}ller]{lecun1998efficient}
Yann LeCun, L{\'e}on Bottou, Genevieve~B. Orr, and Klaus-Robert M{\"u}ller.
\newblock Efficient {BackProp}.
\newblock In \emph{Neural Networks: Tricks of the Trade}, pages 9--50.
  Springer, 1998.

\bibitem[Lee et~al.(2023)Lee, Liu, Bao, Yang, Yu, Cai, Li, Huo, Koutsoukos, and
  Landman]{lee2023repux}
Ho~Hin Lee, Quan Liu, Shunxing Bao, Qi Yang, Xin Yu, Leon~Y. Cai, Thomas~Z. Li,
  Yuankai Huo, Xenofon Koutsoukos, and Bennett~A. Landman.
\newblock Scaling up {3D} kernels with {Bayesian} frequency re-parameterization
  for medical image segmentation.
\newblock In \emph{Medical Image Computing and Computer Assisted Intervention
  (MICCAI)}, pages 632--641, 2023.

\bibitem[Lee et~al.(2026)Lee, Liu, Bao, Huo, and Landman]{lee2026rep3d}
Ho~Hin Lee, Quan Liu, Shunxing Bao, Yuankai Huo, and Bennett~A. Landman.
\newblock Spatially-adaptive gradient re-parameterization for {3D} large kernel
  optimization.
\newblock In \emph{International Conference on Machine Learning (ICML)}, pages
  64654--64672, 2026.

\bibitem[Li et~al.(2021)Li, Kovachki, Azizzadenesheli, Liu, Bhattacharya,
  Stuart, and Anandkumar]{li2021fno}
Zongyi Li, Nikola Kovachki, Kamyar Azizzadenesheli, Burigede Liu, Kaushik
  Bhattacharya, Andrew Stuart, and Anima Anandkumar.
\newblock Fourier neural operator for parametric partial differential
  equations.
\newblock In \emph{International Conference on Learning Representations
  (ICLR)}, 2021.

\bibitem[Lin et~al.(2014)Lin, Maire, Belongie, Hays, Perona, Ramanan,
  Doll{\'a}r, and Zitnick]{lin2014coco}
Tsung-Yi Lin, Michael Maire, Serge Belongie, James Hays, Pietro Perona, Deva
  Ramanan, Piotr Doll{\'a}r, and C.~Lawrence Zitnick.
\newblock {Microsoft COCO}: Common objects in context.
\newblock In \emph{European Conference on Computer Vision (ECCV)}, pages
  740--755, 2014.

\bibitem[Liu et~al.(2023)Liu, Chen, Chen, Chen, Xiao, Wu, K{\"a}rkk{\"a}inen,
  Pechenizkiy, Mocanu, and Wang]{liu2023slak}
Shiwei Liu, Tianlong Chen, Xiaohan Chen, Xuxi Chen, Qiao Xiao, Boqian Wu, Tommi
  K{\"a}rkk{\"a}inen, Mykola Pechenizkiy, Decebal~Constantin Mocanu, and
  Zhangyang Wang.
\newblock More {ConvNets} in the 2020s: Scaling up kernels beyond 51x51 using
  sparsity.
\newblock In \emph{International Conference on Learning Representations
  (ICLR)}, 2023.

\bibitem[Liu et~al.(2022)Liu, Mao, Wu, Feichtenhofer, Darrell, and
  Xie]{liu2022convnext}
Zhuang Liu, Hanzi Mao, Chao-Yuan Wu, Christoph Feichtenhofer, Trevor Darrell,
  and Saining Xie.
\newblock A {ConvNet} for the 2020s.
\newblock In \emph{IEEE/CVF Conference on Computer Vision and Pattern
  Recognition (CVPR)}, pages 11976--11986, 2022.

\bibitem[Loshchilov and Hutter(2019)]{loshchilov2019adamw}
Ilya Loshchilov and Frank Hutter.
\newblock Decoupled weight decay regularization.
\newblock In \emph{International Conference on Learning Representations
  (ICLR)}, 2019.

\bibitem[Luo et~al.(2016)Luo, Li, Urtasun, and Zemel]{luo2016erf}
Wenjie Luo, Yujia Li, Raquel Urtasun, and Richard Zemel.
\newblock Understanding the effective receptive field in deep convolutional
  neural networks.
\newblock In \emph{Advances in Neural Information Processing Systems
  (NeurIPS)}, 2016.

\bibitem[Modoranu et~al.(2026)Modoranu, Safaryan, Schultheis, Riabinin,
  Chumachenko, and Alistarh]{modoranu2026trion}
Ionut-Vlad Modoranu, Mher Safaryan, Erik Schultheis, Maksim Riabinin, Artem
  Chumachenko, and Dan Alistarh.
\newblock Trion: {FFT}-based dynamic subspace selection for low-rank adaptive
  optimization of {LLMs}.
\newblock In \emph{International Conference on Learning Representations
  (ICLR)}, 2026.

\bibitem[Nah et~al.(2017)Nah, Kim, and Lee]{nah2017gopro}
Seungjun Nah, Tae~Hyun Kim, and Kyoung~Mu Lee.
\newblock Deep multi-scale convolutional neural network for dynamic scene
  deblurring.
\newblock In \emph{IEEE Conference on Computer Vision and Pattern Recognition
  (CVPR)}, pages 3883--3891, 2017.

\bibitem[Narayan et~al.(1983)Narayan, Peterson, and
  Narasimha]{narayan1983tdlms}
S. Narayan, A.~M. Peterson, and M.~J. Narasimha.
\newblock Transform domain {LMS} algorithm.
\newblock \emph{IEEE Transactions on Acoustics, Speech, and Signal Processing},
  31\penalty0 (3):\penalty0 609--615, 1983.

\bibitem[Oquab et~al.(2024)Oquab, Darcet, Moutakanni, Vo, Szafraniec, Khalidov,
  Fernandez, Haziza, Massa, El-Nouby, Assran, Ballas, Galuba, Howes, Huang, Li,
  Misra, Rabbat, Sharma, Synnaeve, Xu, J{\'e}gou, Mairal, Labatut, Joulin, and
  Bojanowski]{oquab2024dinov2}
Maxime Oquab, Timoth{\'e}e Darcet, Th{\'e}o Moutakanni, Huy~V. Vo, Marc
  Szafraniec, Vasil Khalidov, Pierre Fernandez, Daniel Haziza, Francisco Massa,
  Alaaeldin El-Nouby, Mahmoud Assran, Nicolas Ballas, Wojciech Galuba, Russell
  Howes, Po-Yao Huang, Shang-Wen Li, Ishan Misra, Michael Rabbat, Vasu Sharma,
  Gabriel Synnaeve, Hu Xu, Herv{\'e} J{\'e}gou, Julien Mairal, Patrick Labatut,
  Armand Joulin, and Piotr Bojanowski.
\newblock {DINOv2}: Learning robust visual features without supervision.
\newblock \emph{Transactions on Machine Learning Research}, 2024.

\bibitem[Radford et~al.(2021)Radford, Kim, Hallacy, Ramesh, Goh, Agarwal,
  Sastry, Askell, Mishkin, Clark, Krueger, and Sutskever]{radford2021clip}
Alec Radford, Jong~Wook Kim, Chris Hallacy, Aditya Ramesh, Gabriel Goh,
  Sandhini Agarwal, Girish Sastry, Amanda Askell, Pamela Mishkin, Jack Clark,
  Gretchen Krueger, and Ilya Sutskever.
\newblock Learning transferable visual models from natural language
  supervision.
\newblock In \emph{International Conference on Machine Learning (ICML)}, pages
  8748--8763, 2021.

\bibitem[Rao et~al.(2021)Rao, Zhao, Zhu, Lu, and Zhou]{rao2021gfnet}
Yongming Rao, Wenliang Zhao, Zheng Zhu, Jiwen Lu, and Jie Zhou.
\newblock Global filter networks for image classification.
\newblock In \emph{Advances in Neural Information Processing Systems
  (NeurIPS)}, pages 980--993, 2021.

\bibitem[Ren et~al.(2015)Ren, He, Girshick, and Sun]{ren2015faster}
Shaoqing Ren, Kaiming He, Ross Girshick, and Jian Sun.
\newblock Faster {R-CNN}: Towards real-time object detection with region
  proposal networks.
\newblock In \emph{Advances in Neural Information Processing Systems
  (NeurIPS)}, 2015.

\bibitem[Rim et~al.(2020)Rim, Lee, Won, and Cho]{rim2020realblur}
Jaesung Rim, Haeyun Lee, Jucheol Won, and Sunghyun Cho.
\newblock Real-world blur dataset for learning and benchmarking deblurring
  algorithms.
\newblock In \emph{European Conference on Computer Vision (ECCV)}, pages
  184--201, 2020.

\bibitem[Rippel et~al.(2015)Rippel, Snoek, and Adams]{rippel2015spectral}
Oren Rippel, Jasper Snoek, and Ryan~P. Adams.
\newblock Spectral representations for convolutional neural networks.
\newblock In \emph{Advances in Neural Information Processing Systems
  (NeurIPS)}, 2015.

\bibitem[Roy et~al.(2023)Roy, Koehler, Ulrich, Baumgartner, Petersen, Isensee,
  J{\"a}ger, and Maier-Hein]{roy2023mednext}
Saikat Roy, Gregor Koehler, Constantin Ulrich, Michael Baumgartner, Jens
  Petersen, Fabian Isensee, Paul~F. J{\"a}ger, and Klaus~H. Maier-Hein.
\newblock {MedNeXt}: Transformer-driven scaling of {ConvNets} for medical image
  segmentation.
\newblock In \emph{Medical Image Computing and Computer Assisted Intervention
  (MICCAI)}, pages 405--415, 2023.

\bibitem[S{\'a}nchez et~al.(1995)S{\'a}nchez, Garc{\'i}a, Peinado, Segura, and
  Rubio]{sanchez1995dct}
V. S{\'a}nchez, P. Garc{\'i}a, A.~M. Peinado, J.~C. Segura, and A.~J. Rubio.
\newblock Diagonalizing properties of the discrete cosine transforms.
\newblock \emph{IEEE Transactions on Signal Processing}, 43\penalty0
  (11):\penalty0 2631--2641, 1995.

\bibitem[Shen et~al.(2019)Shen, Wang, Lu, Shen, Ling, Xu, and
  Shao]{shen2019hide}
Ziyi Shen, Wenguan Wang, Xiankai Lu, Jianbing Shen, Haibin Ling, Tingfa Xu, and
  Ling Shao.
\newblock Human-aware motion deblurring.
\newblock In \emph{IEEE/CVF International Conference on Computer Vision
  (ICCV)}, pages 5571--5580, 2019.

\bibitem[Strang(1999)]{strang1999dct}
Gilbert Strang.
\newblock The discrete cosine transform.
\newblock \emph{SIAM Review}, 41\penalty0 (1):\penalty0 135--147, 1999.

\bibitem[Tieleman and Hinton(2012)]{tieleman2012rmsprop}
Tijmen Tieleman and Geoffrey Hinton.
\newblock Lecture 6.5---{RMSprop}: Divide the gradient by a running average of
  its recent magnitude.
\newblock COURSERA: Neural Networks for Machine Learning, 2012.

\bibitem[Touvron et~al.(2021)Touvron, Cord, Douze, Massa, Sablayrolles, and
  J{\'e}gou]{touvron2021deit}
Hugo Touvron, Matthieu Cord, Matthijs Douze, Francisco Massa, Alexandre
  Sablayrolles, and Herv{\'e} J{\'e}gou.
\newblock Training data-efficient image transformers \& distillation through
  attention.
\newblock In \emph{International Conference on Machine Learning (ICML)}, pages
  10347--10357, 2021.

\bibitem[Ulicny et~al.(2022)Ulicny, Krylov, and Dahyot]{ulicny2022harmonic}
Matej Ulicny, Vladimir~A. Krylov, and Rozenn Dahyot.
\newblock Harmonic convolutional networks based on discrete cosine transform.
\newblock \emph{Pattern Recognition}, 129:\penalty0 108707, 2022.

\bibitem[Vyas et~al.(2025)Vyas, Morwani, Zhao, Shapira, Brandfonbrener, Janson,
  and Kakade]{vyas2024soap}
Nikhil Vyas, Depen Morwani, Rosie Zhao, Itai Shapira, David Brandfonbrener,
  Lucas Janson, and Sham Kakade.
\newblock {SOAP}: Improving and stabilizing {Shampoo} using {Adam} for language
  modeling.
\newblock In \emph{International Conference on Learning Representations
  (ICLR)}, 2025.

\bibitem[Woo et~al.(2023)Woo, Debnath, Hu, Chen, Liu, Kweon, and
  Xie]{woo2023convnextv2}
Sanghyun Woo, Shoubhik Debnath, Ronghang Hu, Xinlei Chen, Zhuang Liu, In~So
  Kweon, and Saining Xie.
\newblock {ConvNeXt V2}: Co-designing and scaling {ConvNets} with masked
  autoencoders.
\newblock In \emph{IEEE/CVF Conference on Computer Vision and Pattern
  Recognition (CVPR)}, pages 16133--16142, 2023.

\bibitem[Xiao et~al.(2018)Xiao, Liu, Zhou, Jiang, and Sun]{xiao2018upernet}
Tete Xiao, Yingcheng Liu, Bolei Zhou, Yuning Jiang, and Jian Sun.
\newblock Unified perceptual parsing for scene understanding.
\newblock In \emph{European Conference on Computer Vision (ECCV)}, pages
  432--448, 2018.

\bibitem[Xie and Li(2024)]{xie2024implicit}
Shuo Xie and Zhiyuan Li.
\newblock Implicit bias of {AdamW}: $\ell_\infty$-norm constrained
  optimization.
\newblock In \emph{International Conference on Machine Learning (ICML)}, pages
  54488--54510, 2024.

\bibitem[Xie et~al.(2025)Xie, Mohamadi, and Li]{xie2025adam}
Shuo Xie, Mohamad~Amin Mohamadi, and Zhiyuan Li.
\newblock Adam exploits $\ell_\infty$-geometry of loss landscape via
  coordinate-wise adaptivity.
\newblock In \emph{International Conference on Learning Representations
  (ICLR)}, 2025.

\bibitem[Yang et~al.(2021)Yang, Hu, Babuschkin, Sidor, Liu, Farhi,
  et~al.]{yang2021mup}
Greg Yang, Edward~J. Hu, Igor Babuschkin, Szymon Sidor, Xiaodong Liu, David
  Farhi, et~al.
\newblock Tuning large neural networks via zero-shot hyperparameter transfer.
\newblock In \emph{Advances in Neural Information Processing Systems
  (NeurIPS)}, pages 17084--17097, 2021.

\bibitem[Ye et~al.(2023)Ye, Pan, Jiang, Lu, Fang, and Xu]{ye2023far}
Yun Ye, Yanjie Pan, Qually Jiang, Ming Lu, Xiaoran Fang, and Beryl Xu.
\newblock Frequency-aware re-parameterization for over-fitting based image
  compression.
\newblock In \emph{IEEE International Conference on Image Processing (ICIP)},
  pages 2310--2314, 2023.

\bibitem[Yong et~al.(2020)Yong, Huang, Hua, and Zhang]{yong2020gc}
Hongwei Yong, Jianqiang Huang, Xiansheng Hua, and Lei Zhang.
\newblock Gradient centralization: A new optimization technique for deep neural
  networks.
\newblock In \emph{European Conference on Computer Vision (ECCV)}, pages
  635--652, 2020.

\bibitem[Yu and Wang(2025)]{yu2025mambaout}
Weihao Yu and Xinchao Wang.
\newblock {MambaOut}: Do we really need {Mamba} for vision?
\newblock In \emph{IEEE/CVF Conference on Computer Vision and Pattern
  Recognition (CVPR)}, pages 4484--4496, 2025.

\bibitem[Yu et~al.(2024)Yu, Si, Zhou, Luo, Zhou, Feng, Yan, and
  Wang]{yu2024metaformer}
Weihao Yu, Chenyang Si, Pan Zhou, Mi Luo, Yichen Zhou, Jiashi Feng, Shuicheng
  Yan, and Xinchao Wang.
\newblock {MetaFormer} baselines for vision.
\newblock \emph{IEEE Transactions on Pattern Analysis and Machine
  Intelligence}, 46\penalty0 (2):\penalty0 896--912, 2024.

\bibitem[Zhang et~al.(2025)Zhang, Maes, Milligan, Jolicoeur-Martineau,
  Mitliagkas, Scieur, Lacoste-Julien, and Guille-Escuret]{zhang2025rotation}
Tianyue~H. Zhang, Lucas Maes, Alan Milligan, Alexia Jolicoeur-Martineau,
  Ioannis Mitliagkas, Damien Scieur, Simon Lacoste-Julien, and Charles
  Guille-Escuret.
\newblock Understanding {Adam} requires better rotation dependent assumptions.
\newblock In \emph{Advances in Neural Information Processing Systems
  (NeurIPS)}, pages 115111--115152, 2025.

\bibitem[Zhang et~al.(2021)Zhang, Zeng, and Zhang]{zhang2021ecbsr}
Xindong Zhang, Hui Zeng, and Lei Zhang.
\newblock Edge-oriented convolution block for real-time super resolution on
  mobile devices.
\newblock In \emph{ACM International Conference on Multimedia (ACM MM)}, pages
  4034--4043, 2021.

\bibitem[Zhou et~al.(2017)Zhou, Zhao, Puig, Fidler, Barriuso, and
  Torralba]{zhou2017ade20k}
Bolei Zhou, Hang Zhao, Xavier Puig, Sanja Fidler, Adela Barriuso, and Antonio
  Torralba.
\newblock Scene parsing through {ADE20K} dataset.
\newblock In \emph{IEEE Conference on Computer Vision and Pattern Recognition
  (CVPR)}, pages 633--641, 2017.

\end{thebibliography}
}

\clearpage
\appendix
\appnumbering
\section{Proofs}
\label{sec:app-proofs}\label{sec:app-notation}

A kernel $\vw\in\R^n$ of one input and one output channel has $n=k^2$ taps, and $\ve=n^{-1/2}\one$ is its mean direction.
An orthonormal basis stores the coefficients $\vc=\mF\vw$, where the rows of $\mF$ are the basis kernels, and the mean coordinate is $\ve^\top\vw=(\mF\ve)^\top\vc$.
An optimizer is coordinate-wise if every coefficient follows the same rule applied to its own gradient history, and scale-free if multiplying this history by $a\neq0$ multiplies the steps by $\sign a$.
For a scalar history $\gamma_1,\dots,\gamma_t$ we write $-\eta R_t$ for the step of a single parameter.

\subsection{Mean amplification}
\label{sec:app-amp}

\begin{proof}[Proof of \cref{prop:coherent}]
The gradient of coefficient $j$ is $(\mF\vg_s)_j=\gamma_s(\mF\ve)_j$, a fixed multiple of the scalar history.
Since the optimizer is scale-free and starts from its zero state, the coefficient steps by $-\eta\,\sign\big((\mF\ve)_j\big)R_t$, or not at all if $(\mF\ve)_j=0$.
The mean coordinate therefore moves by $-\eta R_t\sum_j\lvert(\mF\ve)_j\rvert=-\eta AR_t=A\,\Delta\theta_t$.
\end{proof}

A per-tap learning rate multiplies the step of tap $p$ by a fixed factor $\omega_p$, so the same argument gives $A=\sum_p\omega_p/k$ in the pixel basis.
A random rotation makes $\mF\ve$ uniform on the unit sphere of $\R^n$, whose entries have the mean absolute value $\kappa_n=\Gamma(\frac n2)/\big(\sqrt\pi\,\Gamma(\frac{n+1}2)\big)$, so $\E A=n\kappa_n\approx\sqrt{2/\pi}\,k$.
\Cref{tab:A} lists $A$ for the bases and techniques of \cref{tab:sr}, together with the amplification $\lVert\mF\vd\rVert_1$ of the first DCT tilt $\vd$.
Only the DCT basis leaves both at 1.

\subsection{Output shift}
\label{sec:app-shift}

\begin{proof}[Proof of \cref{prop:centre}]
A change of the kernels moves the output by $\Delta z_o(u)=\sum_{i,p}\Delta w_{oi}(p)\,x_i(u+p)$.
When every shifted input $x_i(\cdot+p)$ has the spatial mean $\mu_i$, as with circular padding, the spatial mean of $\Delta z_o$ is $\sum_i\mu_i\Delta s_{oi}$, where $s_{oi}=\one^\top\vw_{oi}=k\,\ve^\top\vw_{oi}$ is the sum of the kernel's taps.

(i)~By \cref{eq:split}, the common gradient of kernel $(o,i)$ is $\mu_iS_o\one$ with $S_o=\sum_u\delta_o(u)\neq0$, a gradient along $\ve$.
By \cref{prop:coherent}, a sign step on it moves the mean coordinate by $-\eta A\sign(\mu_iS_o)$, so $\Delta s_{oi}=-\eta kA\sign(\mu_iS_o)$; any step orthogonal to $\ve$ leaves $s_{oi}$ unchanged.
Summing gives $\sum_i\mu_i\Delta s_{oi}=-\sign(S_o)\,\eta kA\lVert\vmu\rVert_1$.

(ii)~Training-mode BN subtracts the batch mean of every channel, so adding one constant to $z_o$ in all samples leaves the loss unchanged, and $\sum_{b,u}\delta_o(b,u)=0$.
With $S_o(b)=\sum_u\delta_o(b,u)$, the common gradient summed over the batch is $\sum_b\mu_i^{(b)}S_o(b)=\sum_b\big(\mu_i^{(b)}-\bar\mu_i\big)S_o(b)$.
A kernel change moves the spatial mean of sample $b$'s output by $\sum_i\mu_i^{(b)}\Delta s_{oi}$ and the batch mean, which BN subtracts, by $\sum_i\bar\mu_i\Delta s_{oi}$.
For the steps of (i), the centered output of sample $b$ thus moves by $\sum_i\big(\mu_i^{(b)}-\bar\mu_i\big)\Delta s_{oi}$, at most $\eta kA\lVert\vmu^{(b)}-\bar{\vmu}\rVert_1$.
\end{proof}

Layer normalization over channels subtracts the mean over the channels at each position, so it removes only shifts that all channels share.
Centering the input by a running mean $\hat{\vmu}$ removes the common gradient only up to $\vmu-\hat{\vmu}$, and at zero-padded borders the padded taps see $-\hat\mu_i$.

\Cref{eq:taulaw} applies (i) to the first step that exceeds the tolerance $T$.
For Adam the step of (i) is multiplied by $\lvert R_t\rvert$, which is the same in two runs that see the same coherent gradients.
The ratio of \cref{eq:scaling} is thus unchanged, and in \cref{eq:taulaw} $T$ is divided by the largest $\lvert R_t\rvert$ (1 for sign steps).

\begin{table}[t]
\centering
\caption{\textbf{Amplification of the mean ($A$) and of the first DCT tilt ($A_1$)} for the bases and per-tap learning rates of \cref{tab:sr}.}
\label{tab:A}
\scriptsize
\setlength{\tabcolsep}{2pt}
\begin{tabular*}{\linewidth}{@{}l@{\extracolsep{\fill}}c ccc ccc@{}}
\toprule
& & \multicolumn{3}{c}{$A$} & \multicolumn{3}{c}{$A_1$} \\
\cmidrule(lr){3-5}\cmidrule(l){6-8}
Basis or technique & $A$ & $k{=}9$ & 17 & 31 & 9 & 17 & 31 \\
\midrule
\rowcolor{pixelband}Pixel basis & $k$ & 9 & 17 & 31 & 8.02 & 15.3 & 27.9 \\
Random rotation & $\approx0.8k$ & 7.22 & 13.7 & 24.8 & 7.53 & 13.6 & 24.7 \\
Center $3{\times}3$ branch & $k{+}9/k$ & 10.0 & 17.5 & 31.3 & 8.34 & 15.4 & 27.9 \\
Fixed distance mask & $\sum_p\omega_p/k$ & 2.30 & 2.67 & 2.94 & 1.77 & 2.02 & 2.18 \\
Per-layer rate $3/k$ & 3 & 3 & 3 & 3 & 2.67 & 2.70 & 2.70 \\
Uniform rate $1/k$ & 1 & 1 & 1 & 1 & 0.89 & 0.90 & 0.90 \\
Mean-isolating, pixel-like & 1 & 1 & 1 & 1 & 8.02 & 15.3 & 27.9 \\
Mean-isolating, random & 1 & 1 & 1 & 1 & 7.04 & 13.5 & 24.5 \\
\rowcolor{dctband}DCT basis & 1 & 1 & 1 & 1 & 1 & 1 & 1 \\
\bottomrule
\end{tabular*}
\vspace{-0.1in}
\end{table}

\subsection{Optimizers that ignore the basis}
\label{sec:app-equiv}

\begin{proposition}\label{prop:sgd}
From the same kernels and settings, SGD with momentum, weight decay and gradient-norm clipping gives the same deployed kernels in every orthonormal basis.
So do Muon~\cite{jordan2024muon} and Shampoo~\cite{gupta2018shampoo} applied to each layer's kernels as one matrix with one row per output channel, with one basis for all kernels of the layer.
\end{proposition}

\begin{proof}
A change of basis applies an orthogonal $\mQ$ to all coefficients of a layer.
If an optimizer's step from the transformed history of parameters and gradients is $\mQ$ times its step from the original history, then by induction over the steps the coefficients stay $\mQ$ times the original ones, and the deployed kernels coincide.
SGD's momentum, weight decay and learning-rate schedule are linear in the parameters and gradients, and gradient-norm clipping depends only on $\lVert\vg\rVert_2=\lVert\mQ\vg\rVert_2$.
On the matrix view, the change of basis multiplies the gradient matrix $\mX$ from the right by an orthogonal matrix, $\mX\mapsto\mX\mQ$.
Muon's Newton--Schulz iteration is a polynomial in $\mX\mX^\top$ applied to $\mX$, and Shampoo's preconditioners transform as $\mL\mapsto\mL$ and $\mR\mapsto\mQ^\top\mR\mQ$, so both steps transform in the same way.
\end{proof}

\begin{proposition}\label{cor:eps}
Let Adam with $\epsilon>0$ and without weight decay train a kernel from its zero state, with gradients $\vg_s$, $G_t=\max_{s\le t}\lVert\vg_s\rVert_2$ and bias-corrected first moment $\hat{\vm}_t$.
In every orthonormal basis, its deployed step lies within $\frac{G_t}{G_t+\epsilon}\lVert\Delta\vw^{\mathrm{lin}}_t\rVert_2$ of the momentum step $\Delta\vw^{\mathrm{lin}}_t=-(\eta/\epsilon)\hat{\vm}_t$, which does not depend on the basis.
\end{proposition}

\begin{proof}
The second moment of coefficient $j$ is a weighted average of $(\mF\vg_s)_j^2\le\lVert\vg_s\rVert_2^2$, so $\sqrt{\hat v_{t,j}}\le G_t$.
With $\delta_j=\sqrt{\hat v_{t,j}}/\epsilon$, the step of coefficient $j$ is $(\mF\Delta\vw^{\mathrm{lin}}_t)_j/(1+\delta_j)$, which differs from the linear step by the fraction $\delta_j/(1+\delta_j)\le G_t/(G_t+\epsilon)$.
Summing the squares and transforming back with $\mF^\top$ gives the bound.
\end{proof}

When the gradients of a kernel stay far below $\epsilon$, Adam thus takes the basis-independent steps $-(\eta/\epsilon)\hat{\vm}_t$, the $\epsilon$-dominated regime of \cref{sec:equiv}.

\subsection{Per-tap learning rates}
\label{sec:app-diagonal}\label{sec:diagonal}

A small-kernel branch without normalization and a fixed mask act as per-tap learning rates.
Let the deployed kernel be $\vw=\vw_0+\sum_b\mE_b\vw_b$, where each embedding $\mE_b$ places a branch on a set $S_b$ of taps, and let all tensors train with one coordinate-wise optimizer from their zero states.
Every copy of tap $p$ then receives the same gradient history as the tap of the deployed kernel, so all copies take the same step, and the deployed tap moves $r_p=1+\#\{b:p\in S_b\}$ times as far.
A parallel $3\times3$ branch thus doubles the rate of the nine center taps.
A fixed mask $\vw=\vm\odot\vu$ gives scale-free optimizers the per-tap rates $\vm$, up to a rescaled $\epsilon$.

\begin{proposition}\label{thm:mask}
Let the gradient of a kernel be $\gamma\ve+\vxi$ with a fixed $\gamma$ and a centered Gaussian pattern part $\vxi\perp\ve$ of variance $s^2>0$ at every tap, and let $r=\gamma^2/\E\lVert\vxi\rVert^2$.
For sign steps, all fixed per-tap learning rates with $A=1$ make the same expected progress on the pattern, $-\E[\vxi^\top\Delta\vw]=\eta ks\sqrt{2/\pi}\,e^{-r/2}$, and every mean-isolating basis makes at least $e^{r/2}$ times as much.
For steps divided by the root-mean-square gradient of each coefficient, the progress is $\eta ks/\sqrt{1+r}$ and the factor is at least $\sqrt{1+r}$.
\end{proposition}

\begin{proof}
With per-tap rates $\vomega$, tap $p$ steps by $-\eta\omega_p\sign(\gamma/k+\xi_p)$.
For $Z\sim\mathcal N(0,s^2)$, $\E[Z\sign(a+Z)]=2s\varphi(a/s)$, and with $a=\gamma/k$, $a^2/s^2=r$; summing over the taps with $\sum_p\omega_p=kA=k$ gives $\eta ks\sqrt{2/\pi}\,e^{-r/2}$.
In a basis whose first row is $\ve$, the pattern coefficients receive only $\vf_j^\top\vxi$, of standard deviation $s_j$, and their sign steps make the progress $\eta\sqrt{2/\pi}\sum_js_j\ge\eta\sqrt{2/\pi}\,\big(\sum_js_j^2\big)^{1/2}=\eta\sqrt{2/\pi}\,ks$.
The root-mean-square case divides by $(a^2+s^2)^{1/2}$ per tap and by $s_j$ per pattern coefficient and is linear.
\end{proof}

\subsection{The DCT basis}
\label{sec:app-dctopt}\label{sec:app-dct}

Let $\mD$ be the orthonormal one-dimensional DCT-II with rows $\vd_0,\dots,\vd_{k-1}$, and $\mD_2=\mD\otimes\mD$ the separable two-dimensional DCT with rows $\vphi_{ab}=\vd_a\otimes\vd_b$, so $\vphi_{00}=\ve$.
{A $k^2\times k^2$ matrix is \emph{stationary with mirrored boundaries} if it is a linear combination of products $\mX\otimes\mY$ of one-dimensional matrices $a(i-j)+a(i+j+1)$ with $a$ even and $2k$-periodic, that is, of stationary statistics on the window mirrored at its edges.
The DCT-II diagonalizes each such one-dimensional matrix~\cite{sanchez1995dct,strang1999dct}, whose eigenvalues $\sum_la(l)\cos(\pi ml/k)$ take any values as $a$ varies, so these are exactly the matrices $\mD_2^\top\diag(\mTheta)\mD_2$ with $\mTheta\in\R^{k\times k}$.}

\begin{lemma}\label{thm:dct}
The DCT-II diagonalizes every matrix $a(i-j)+a(i+j+1)$ with $a$ even and $2k$-periodic.
\end{lemma}

\begin{proof}
Extend $\vx\in\R^k$ to $\R^{2k}$ by half-sample symmetry, $\tilde x_j=\tilde x_{2k-1-j}=x_j$.
The matrix acts on $\vx$ as the circular convolution of $\tilde{\vx}$ with the even kernel $a$, restricted to the first $k$ entries.
The symmetric extension of $\vd_m$ is a combination of $e^{\pm i\pi mj/k}$, both eigenvectors of this convolution with the eigenvalue $\sum_la(l)\cos(\pi ml/k)$, so $\vd_m$ is an eigenvector.
\end{proof}

\begin{proof}[Proof of \cref{prop:dctbasis}]
Every $\mSigma=\mD_2^\top\diag(\mTheta)\mD_2$ is diagonal in the DCT basis.
If $\mTheta$ has distinct entries, such as $\Theta_{ab}=1+ka+b$, the eigendirections of $\mSigma$ are exactly $\pm\vphi_{ab}$, so an orthonormal basis contains all of them, or equivalently diagonalizes $\mSigma$, if and only if it is the DCT basis up to sign flips and reordering.
Since the class contains such matrices, a basis that diagonalizes every matrix of the class is the DCT basis up to sign flips and reordering.
This basis contains $\ve=\vphi_{00}$, and for every row $\vf$ of an orthonormal basis $\lVert\mF\vf\rVert_1=1$, so with distinct eigenvalues it amplifies neither the mean nor any other eigendirection.

For the quadratic part we use exact gradients, no weight decay, zero initial optimizer state, $\sign(0)=0$ and a zero update for a coefficient whose gradient history is zero.
For the quadratic loss $\frac12(\vw-\vw^\star)^\top\mH(\vw-\vw^\star)$ with $\mD_2\mH\mD_2^\top=\diag(\vtheta)$, $\vtheta>0$, coefficient $j$ receives the gradient $\theta_j(c_j-c^\star_j)$.
As long as the trajectory agrees with that for $\mH=\mI$, this history is $\theta_j$ times the one for $\mH=\mI$, and a scale-free optimizer takes the same step.
By induction the two trajectories coincide.
Conversely, from the zero state an optimizer with $\epsilon=0$ such as Adam, RMSprop, signSGD or Lion first steps by $-\eta\sigma_0\sign(\vg)$ with $\sigma_0>0$.
Equal first steps for $\mH$ and $\mI$ in a basis $\mF$ require $\sign(\mF\mH\mF^\top\vz)=\sign\vz$ for every $\vz=\mF(\vw_0-\vw^\star)$, and $\vz$ equal to a unit vector $\mathbf{e}_j$ forces every off-diagonal entry of column $j$ of $\mF\mH\mF^\top$ to vanish, so $\mF\mH\mF^\top$ is diagonal.
A basis with this property for every such Hessian thus diagonalizes one with distinct eigenvalues, so it is the DCT basis up to sign flips and reordering.
\end{proof}

The DCT diagonalizes stationary statistics without mirrored boundaries up to a relative off-diagonal part that vanishes as $k$ grows.
For a symmetric Toeplitz matrix $\mT=[t_{\lvert i-j\rvert}]$, setting $a(l)=t_l$ for $l<k$, $a(k)=0$ and $a(l)=t_{2k-l}$ for $l>k$ writes $\mT$ as the matrix of \cref{thm:dct} minus a Hankel matrix $[a(i+j+1)]$.
The part of $\mD\mT\mD^\top$ off the diagonal therefore has a squared Frobenius norm of at most $2\sum_{l=1}^{k-1}l\,t_l^2$, a fraction $O(1/k)$ of $\lVert\mT\rVert_F^2$ when $\sum_ll\,t_l^2<\infty$.
For the condition number in \cref{sec:method}, we model Adam with frozen second moments by the preconditioner $\mF^\top\diag(\mF\mSigma\mF^\top)^{-1/2}\mF$ with curvature $\mH=\mSigma$.
$\mSigma$ is the covariance of 2,000 luminance patches from each of 200 $\times2$ low-resolution DIV2K images, centered and trace-normalized, plus the mean component, with a ridge of $10^{-6}$.
For this covariance, the condition number in the DCT basis is within 3--8\% of the eigenbasis at $k=3$, 9, 17 and 31.

The same holds for the kernel gradients of trained networks.
We measure their second moment $\mS=\E[\vg\vg^\top]$ in the SR network at $k=7$--31 and in {the first two stages of} ConvNeXt-T with 31--29--27--13 kernels trained on ImageNet-100{, whose kernels are at most about as wide as their feature maps}.
The DCT basis leaves 3--19\% of the pattern energy off the diagonal, against 43--93\% for the Householder and random-pattern bases, and less than both in 99--100\% of the SR network's kernels.
With a frozen normalizer and curvature $\mS$, its condition number is 1.2--2.7 times that of the eigenbasis of $\mS$, whereas the pixel basis is 8.5--66 times worse conditioned than the DCT basis.

\section{Protocols}
\label{sec:app-protocols}

\subsection{The SR network}
\label{sec:app-sr}

The SR network of \cref{sec:validation} follows ECBSR~\cite{zhang2021ecbsr} for $\times2$ super-resolution of the luminance.
Its input is the Y channel in $[0,255]$.
It trains on DIV2K~\cite{agustsson2017div2k} with Adam and the L1 loss on batches of 32 patches of $64^2$, and a basis applies to every convolution.
For \cref{tab:sr,tab:reparam} and \cref{fig:gain,fig:pred}b, each configuration trains for 300k updates with cosine decay at the rate, on a $\sqrt2$ grid, that gives the best final PSNR on all 100 DIV2K validation images.
Each entry is the mean of three runs at this rate.
At $k=31$ every configuration trains on $128^2$ patches in batches of 16, since a $64^2$ patch is only about twice as wide as the kernel.
We report the Urban100 Y-PSNR (BT.601, 2-pixel border) of the folded network.

\textbf{Maximal learning rate.}
To measure $\tau$, a sweep trains one network per rate, on a factor-2 grid with $\sqrt2$ points near the threshold, for 16k updates.
Counting upwards from the best rate of the sweep, $\tau$ is the largest rate up to which every run ends within 0.5\,dB of the best validation PSNR.

\textbf{Interventions.}
The mean-isolating basis with pixel-like pattern rows is the Householder reflection that maps the first pixel axis to $\ve$.
In \cref{tab:norm}, BN uses batch statistics in training (momentum 0.1) and is folded into the kernel at export, LN normalizes over the channels at each position, and centering subtracts a running mean of the input (momentum 0.1).
The $3\times3$ branch is added at the center of every body kernel, with BN on both paths in its BN variant.
In \cref{fig:pred}b, each optimizer selects its rate in each basis, and SGD one rate for both bases.
{The per-layer rate $3/k$ and the uniform rate $1/k$ of \cref{tab:sr} scale the learning rate of the four body kernels only, and every other parameter keeps the base rate.}

\textbf{Structural re-parameterizations.}
In \cref{tab:reparam} every method changes only the four body convolutions and folds exactly into the same deployed network.
The branch methods follow their original blocks with BN on every path: ACNet adds $k\times1$ and $1\times k$ kernels~\cite{ding2019acnet}, the RepVGG-style block a $1\times1$ branch and an identity~\cite{ding2021repvgg}, DBB its four branches~\cite{ding2021dbb}, RepLKNet and SLaK a parallel small kernel~\cite{ding2022replknet,liu2023slak}, here $3\times3$ and in their networks $5\times5$, and UniRepLKNet its dilated kernels~\cite{ding2024unireplknet}.
ECB adds its branches without BN~\cite{zhang2021ecbsr}, and Rep3D generates a mask from the distance mask with a small network~\cite{lee2026rep3d}.
The larger expansion trains every convolution as 15--17 parallel branches of $1\times1$--$k\times k$--$1\times1$ chains with $4\times$ expansion.

\subsection{Standard networks}
\label{sec:app-standard}

On every standard network only the basis of the large kernels changes, and both bases use the same recipe, with the learning rate selected as described below.
Every network is evaluated after its last update, folded into the original architecture.
For each test set, we report the share of images on which \ours{} reaches a higher PSNR.

\textbf{NAFNet.}
We enlarge every $3\times3$ depthwise kernel of the width-32 NAFNet~\cite{chen2022nafnet} to $k\times k$ and keep its official GoPro recipe.
It trains on the 2,103 GoPro training images~\cite{nah2017gopro} and is tested on GoPro and HIDE~\cite{shen2019hide} with TLC~\cite{chu2022tlc}.

\textbf{ConvNeXt-T and dense prediction.}
ConvNeXt-T~\cite{liu2022convnext} with 31--29--27--13 kernels trains on ImageNet-1K with
SLaK's 120-epoch recipe~\cite{liu2023slak}, without BN after the depthwise convolutions, as in ConvNeXt.
Each basis selects its base rate by the top-1 accuracy on 10 training images per class held out from training.
Detection and segmentation fine-tune the two backbones trained at $4\times10^{-3}$ with SLaK's protocols: UperNet~\cite{xiao2018upernet} on ADE20K, Faster R-CNN~\cite{ren2015faster} on VOC 2007+2012, and Cascade Mask R-CNN~\cite{cai2018cascade} on COCO.

\textbf{ImageNet-100.}
ImageNet-100 consists of 100 ImageNet classes, and the official validation images of these classes serve as the test set.
For the gradient measurements, ConvNeXt-T with 31--29--27--13 kernels trains on it for 100 epochs, 20 of them warm-up, with AdamW (weight decay 0.05) and a global batch of 1,024.
For \cref{fig:in100law}, ConvNeXt-T trains at $256\times256$, so that its four stages see feature maps of 64, 32, 16 and 8 pixels.
Its depthwise kernels are $k\times k$ in the first two stages and $\min(k,S-1)$ in the later ones, where $S$ is the stage's feature-map size.
Each run lasts 12 epochs, one of linear warm-up and eleven at a constant rate{, and in both bases the large kernels train at four times the base rate of $4\times10^{-3}$ and all other weights at the base rate}.

\textbf{ViT-S/16.}
ViT-S/16 trains with the 90-epoch recipe of Beyer~\etal~\cite{beyer2022better}.
\ours{} stores each $16\times16$ slice of the patch embedding as DCT coefficients.
Each selects its peak rate on 10 held-out images per class.

\textbf{Released models in \cref{tab:probe}.}
Each network receives 256 ImageNet validation images in 8 batches of 32, in training mode, and we back-propagate the cross-entropy of its pretrained head (a random head for self-supervised models).
From the 8 gradients of a kernel we form $\vm$ and $\vv$, the means of $\vg_b$ and $\vg_b^2$, in the pixel and the DCT basis and the Adam-like step $\vm/(\sqrt{\vv}+10^{-8})$.
A layer's ratio is the root-mean-square mean step in the pixel basis over that in the DCT basis, averaged over the probed layers, for UniRepLKNet-T its 15 $13\times13$ layers.

\section{Additional results}
\label{sec:app-results}\label{sec:app-lowlevel}

\textbf{Deblurring.}
On RealBlur-J~\cite{rim2020realblur}, which has no validation split, each basis selects its learning rate on 84 of its training pairs.
NAFNet-k13 then gains 0.40\,dB (aligned PSNR with TLC) and is better on 78\% of the 980 test images.

\begin{table}[t]
\centering
\caption{\textbf{\ours{} is more accurate at every learning rate.} ImageNet-1K top-1 accuracy (\%) at each learning rate ($10^{-3}$).}
\label{tab:app-cls}\label{tab:app-in1k}\label{tab:vit}
\scriptsize
\setlength{\tabcolsep}{2pt}
\begin{tabular*}{\linewidth}{@{}l@{\extracolsep{\fill}}cccc@{}}
\toprule
ConvNeXt-T, base rate & 2 & 4 & 8 & 16 \\
\midrule
\rowcolor{pixelband}Baseline & 79.80 & 80.43 & 80.50 & 79.78 \\
\rowcolor{dctband}\ours{} & 80.23 & 80.96 & 81.22 & 80.91 \\
\midrule
ViT-S/16, peak rate & 1 & 2 & 4 & \\
\midrule
\cellcolor{pixelband}Baseline & \cellcolor{pixelband}76.71 & \cellcolor{pixelband}76.61 & \cellcolor{pixelband}75.26 & \\
\cellcolor{dctband}\ours{} (patch embedding) & \cellcolor{dctband}77.09 & \cellcolor{dctband}77.14 & \cellcolor{dctband}75.74 & \\
\bottomrule
\end{tabular*}
\vspace{0.12in}
\centering
\caption{\textbf{Training cost relative to the plain convolution}: median step time, trainable parameters and peak memory.}
\label{tab:app-cost}
\scriptsize
\setlength{\tabcolsep}{2pt}
\begin{tabular*}{\linewidth}{@{}ll@{\extracolsep{\fill}}ccc@{}}
\toprule
Setting & Method & Time & Param. & Memory \\
\midrule
SR network, & \ours{} & 1.01\,/\,1.06 & 1.00 & 1.00\,/\,1.02 \\
$k=9$\,/\,17 & $3\times3$ branch, no BN & 1.03\,/\,1.04 & 1.11\,/\,1.03 & 1.00\,/\,1.01 \\
 & Fixed distance mask & 1.02\,/\,1.03 & 1.00 & 1.00\,/\,1.01 \\
 & Learned mask & 1.39\,/\,1.22 & 1.00 & 1.00\,/\,1.01 \\
 & Larger expansion & 1.71\,/\,1.47 & 54.8\,/\,49.9 & 1.54\,/\,2.73 \\
\midrule
ConvNeXt-T & \ours{} & 1.00 & 1.00 & 1.00 \\
NAFNet-k13 & \ours{} & 1.00 & 1.00 & 1.00 \\
\bottomrule
\end{tabular*}
\vspace{-0.1in}
\end{table}

\textbf{Classification.}
\Cref{tab:app-cls} gives the classification results at every rate.
The held-out images select $4\times10^{-3}$ for the baseline and $8\times10^{-3}$ for \ours{} on ConvNeXt-T, and $10^{-3}$ and $2\times10^{-3}$ on ViT-S/16.

\begin{figure}[t]
\centering
\includegraphics{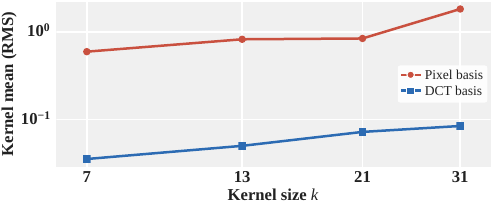}
\vspace{-0.35in}
\caption{{\textbf{On ImageNet, the pixel basis lets the kernel means drift.}} ConvNeXt-T on ImageNet-100 after 12 epochs at constant learning rates. {Root mean square of the first-stage kernels' means when both bases train the large kernels at four times the base rate.}}
\label{fig:in100law}
\end{figure}

\textbf{Mean amplification on ImageNet-100.}
In ConvNeXt-T with 31--29--27--13 kernels trained on ImageNet-100 in either basis, the mean direction carries 25--28\%, 18--19\% and 7--9\% of the energy of the kernel gradients in stage 1, stage 2 and stages 3--4, against at most 0.6\% for a random direction.
When both bases train the large kernels at the same learning rate, {the first-stage kernel means} end 12 to 22 times larger in the pixel basis than in the DCT basis (\cref{fig:in100law}).

\textbf{Cost and code.}
\ours{} keeps the number of parameters and the optimizer state and forms each kernel by two small matrix products per step.
\Cref{tab:app-cost} gives the measured cost, and \cref{fig:app-code} a drop-in module.
Each method and the plain convolution run in one process on one AMD Instinct MI325X GPU and are timed in alternating blocks after warm-up.
A step time is the median over the blocks of the ratio to the plain convolution; \cref{tab:reparam} uses the same measurements.

\begin{figure}[t]
\centering
\scriptsize
\newsavebox{\codebox}\setlength{\topsep}{0pt}\setlength{\partopsep}{0pt}
\begin{lrbox}{\codebox}\begin{minipage}{\dimexpr\linewidth-2\fboxsep-2\fboxrule\relax}
\begin{verbatim}
import math, torch, torch.nn as nn

def dct_matrix(k):  # orthonormal: D @ D.T = I
    x = torch.arange(k, dtype=torch.float64)
    D = torch.cos(math.pi*x[:, None]*(2*x+1)/(2*k))
    D[0] /= math.sqrt(2)
    return D * math.sqrt(2 / k)

class DCTConv2d(nn.Module):
    """A k x k conv trained in the DCT basis."""
    def __init__(self, conv: nn.Conv2d):
        super().__init__()
        self.conv = conv  # keeps stride, padding
        k = conv.kernel_size[0]
        D = dct_matrix(k).to(conv.weight)
        self.register_buffer("D", D)
        W = conv.weight.detach()
        self.coef = nn.Parameter(D @ W @ D.T)
        del conv.weight  # optimizer sees coef only
    def kernel(self):
        return self.D.T @ self.coef @ self.D
    def forward(self, x):  # any padding mode
        return self.conv._conv_forward(
            x, self.kernel(), self.conv.bias)
    @torch.no_grad()
    def fold(self):  # once, after training
        w = nn.Parameter(self.kernel())
        self.conv.weight = w
        return self.conv
\end{verbatim}
\end{minipage}\end{lrbox}
\fbox{\usebox{\codebox}}
\vspace{-0.1in}
\caption{\textbf{\ours{} as a drop-in module.} Wrap each large square convolution before creating the optimizer, train with the unchanged recipe, and call \texttt{fold()} once to obtain a plain \texttt{nn.Conv2d}.}
\label{fig:app-code}
\vspace{-0.1in}
\end{figure}

\end{document}